\documentclass{article}

\usepackage{cite,amsmath,graphicx}
\usepackage[caption=false,font=footnotesize]{subfig}
\usepackage{bm}
\usepackage{amssymb}
\usepackage{booktabs}
\usepackage{multirow}
\usepackage{algorithm,algorithmic}

\usepackage{epstopdf}
\usepackage{amsthm}
\usepackage{lineno}
\usepackage{mathrsfs}
\usepackage{amsmath}
\usepackage{graphicx}
\usepackage{caption}
\usepackage{cases}
\usepackage{xcolor}

\newtheorem{definition}{Definition}

\DeclareMathOperator*{\argmin}{argmin}

\begin{document}
%
\title{Adaptively Topological Tensor Network for Multi-view Subspace Clustering}

\author{Yipeng~Liu, Yingcong~Lu, Weiting Ou, Zhen Long, Ce~Zhu
\thanks{All the authors are with the School of Information and Communication Engineering, University of Electronic Science and Technology of China (UESTC), Chengdu, 611731, China. E-mail: yipengliu@uestc.edu.cn}
}

\maketitle

\begin{abstract}
Multi-view subspace clustering methods have employed learned self-representation tensors from different tensor decompositions to exploit low rank information. However, the data structures embedded with self-representation tensors may vary in different multi-view datasets. Therefore, a pre-defined tensor decomposition may not fully exploit low rank information for a certain dataset, resulting in sub-optimal multi-view clustering performance. To alleviate the aforementioned limitations, we propose the adaptively topological tensor network (ATTN) by determining the edge ranks from the structural information of the self-representation tensor, and it can give a better tensor representation with the data-driven strategy. Specifically, in multi-view tensor clustering, we analyze the higher-order correlations among different modes of a self-representation tensor, and prune the links of the weakly correlated ones from a fully connected tensor network. Therefore, the newly obtained tensor networks can efficiently explore the essential clustering information with  self-representation with different tensor structures for various datasets. A greedy adaptive rank-increasing strategy is further applied to improve the capture capacity of low rank structure. We apply ATTN on multi-view subspace clustering and utilize the alternating direction method of multipliers to solve it. Experimental results show that multi-view subspace clustering based on ATTN outperforms the counterparts on six multi-view datasets.
\end{abstract}

\maketitle

\section{Introduction}
\label{Introduction}
Clustering is an important machine learning technique with applications in multiple fields, such as computer vision, data mining, biomedical engineering \cite{wang2021fast, liu2015matching, liu2022tensor}. It aims to divide data samples into different groups according to their similarities. Owing to increasingly affordable feature collectors, the object can be described from different views. For instance, a certain event is reported by diverse media organizations with different pieces of news; multiple videos are acquired from different kinds of cameras for the same scenario; an image can be described by various types of features \cite{yang2018multi}. These multi-view data provide complementary and consistent information simultaneously to boost the clustering performance~\cite{kang2019low, huang2019multi}. A number of multi-view subspace clustering (MSC) models have been proposed to fully exploit essential information~\cite{chen2022multi,liu2021multiview,sun2021scalable}. 

Most MSC methods are conducted by generating a latent subspace representation and feeding it into the spectral clustering algorithm \cite{wang2021fast}. Matrix or tensor decomposition serves as one of the key techniques for subspace representation \cite{liu2022tensor}. Works in \cite{liu2013multi, brbic2018multi,gao2019multi} have employed the non-negative matrix factorization (NMF) to leverage the multi-view data. A novel bilinear factorization proposed by Zheng et al. \cite{zheng2020constrained} aims to obtain the underlying consensus information across different views effectively. Since the matrix representation based MSC methods always treat the view-specific data matrices separately and lack of capacity to capture inter-view information, we mainly focus on the tensor based multi-view subspace clustering (TMSC) approaches which stack the multiple self-representation matrices into a 3rd-order tensor. 
By employing different low rank tensor constraints on the obtained representation tensor, TMSC has advantages over capturing the high-order information across different views, leading to promising performance~\cite{jia2021multi,chen2022low}. As for the tensor decomposition, the CANDECOMP/PARAFAC (CP) decomposition and Tucker decomposition are two classical methods, which have been applied in TMSC models to discover the low rank information embedded in a self-representation tensor~\cite{yang2016learning,chen2021low,lu2021multi}. However, CP rank is NP-hard to compute generally, which leads to ill-posedness and computational issues. Furthermore, amounts of previous work show that in high-order tensor decomposition, CP and Tucker decompositions tend to degrade rapidly, failing to fully exploit the high-order information of self-representation tensor~\cite{yu2022online}. 
Because of convexity and the good capability of exploiting the low rank information of 3rd-order tensor, the tensor singular value decomposition based tensor nuclear norm (t-SVD-TNN) has been introduced for TMSC~\cite{sun2020multi, li2022high, wang2022hyper}. Although t-SVD based MSC method rotates the self-representation tensor to explore the inter-view correlations, the high-dimensional correlation information has yet to be fully mined in high order tensor because the t-SVD operation can't capture the inter-views and intra-view information simultaneously. Owing to the incremental data volume, the development of tensor network models has been spurred \cite{liu2022tensor}. To capture the local feature and geometric information of tensor data, a graph regularized non-negative tensor ring (GNTR) decomposition is used for MSC \cite{yu2022graph}. Compared with the state-of-the-art tensor based methods in clustering tasks, it achieves better clustering results. In addition, He et al. \cite{wang2022hyper} employ the TR decomposition and generate the multi-mode tensor subspace clustering algorithm (MMTSC) for the tensor space clustering with or without missing entries. However, compared to a general tensor network, the tensor ring only establishes a connection between two factors, leading to a limited characterization for correlations of high-order tensors.

Generally, the aforementioned TMSC methods can obtain competitive clustering performance in some applications, but there are still some challenges. Firstly, the topological architecture and the factors calculation of the tensor decomposition involved in the existing models will cause problems. For example, Tucker decomposition based tensor nuclear norm is simply a rank-sum norm, whose concerning matrices is obtained by matricizing the tensor along one single mode and thus is highly unbalanced~\cite{bengua2017efficient}.
Moreover, it describes the low rank information in the matrix-SVD based space, leading to the sub-optimal representation~\cite{xie2018unifying}.
As for the t-SVD, it suffers from rotation sensitivity and has defects in mining the information along the 3rd mode of the self-representation tensor \cite{liu2018improved}, which makes the t-SVD based TMSC methods can not capture the intra-view and inter-views information simultaneously. Since the TR decomposition only considers the correlations of adjacent modes and has a high sensitivity to the permutation of tensor modes \cite{zheng2021fully}, the resultant self-representation tensor may be inflexible and inadequate in high-order cases. Secondly, the tensor decomposition based MSC methods need the ranks to be pre-determined. As in \cite{chen2021low}, multiple rank-fitting experiments have been additionally employed to obtain the optimal ranks of different representation tensors, which is inefficient and troublesome for dealing with the multi-view clustering task. Thirdly, multi-view datasets come from various application fields, whose self-representation tensors may have different low rank structures for optimal clustering. This implies that capturing the essential clustering information by certain tensor constraints may lead to sub-optimal subspace representation and clustering performance.

To remedy the above problems, we propose an adaptively topological tensor network (ATTN) framework to search for a nearly optimal tensor topology from a self-representation tensor, and apply it to solve TMSC tasks. The proposed model can adaptively find out the topological network architecture that almost best fits its own structure from the corresponding self-representation tensors based on original multi-view data. Instead of simply increasing the rank of the network from 1 to search for a suitable network structure, our algorithm uses correlation information to search for a suitable network structure by removing redundant connections from the fully connected tensor network (FCTN) \cite{zheng2021fully}. After that, we propose the rank increment algorithm based on the greedy algorithm to automatically learn the optimal rank of the network, which improves its representation ability and captures the relevant intra-view and inter-views information better. Experimental results on six multi-view datasets demonstrate that the proposed data-driven tensor network outperforms existing methods on six clustering metrics in clustering tasks.

\begin{figure*}[htbp]
\centering
\centering
\includegraphics[scale=0.35]{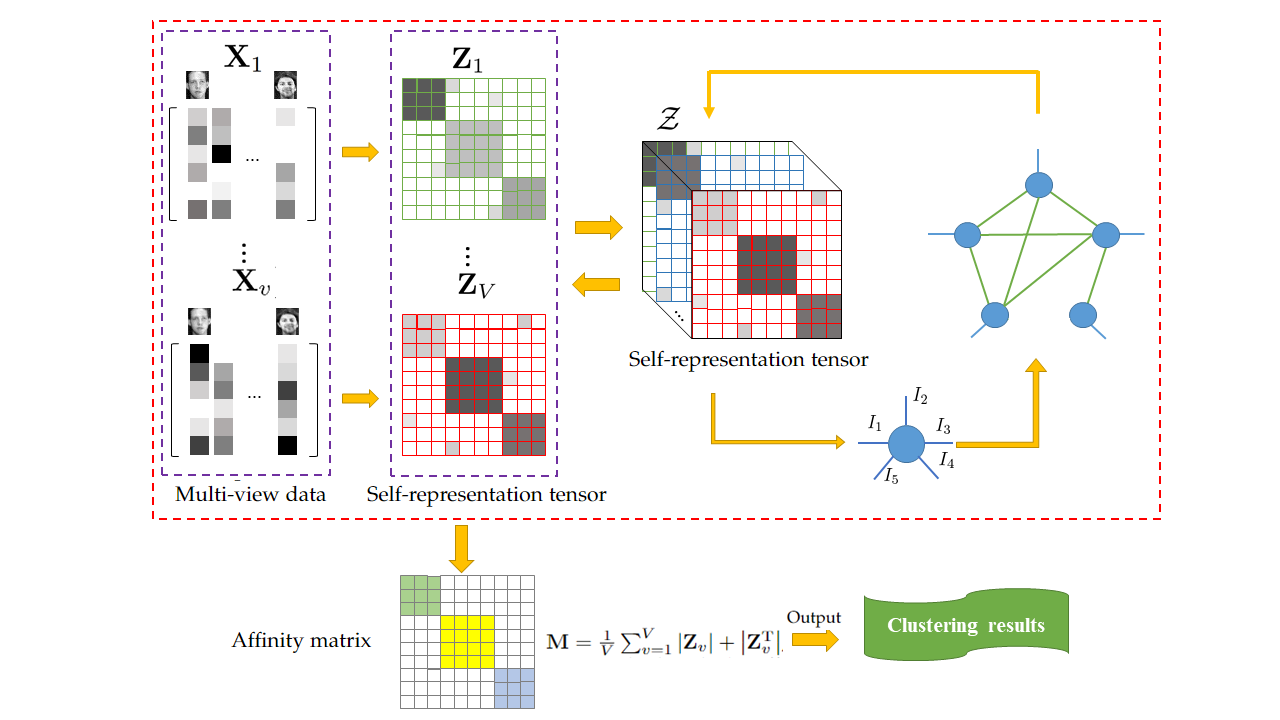}
\caption{The pipeline of our proposed ATTN-MSC. Given multi-view dataset $\{\mathbf{X}_{v}\},v=1,\cdots,V$, we obtain their relevant self-representation matrices $\{\mathbf{X}_{v}\},v=1,\cdots,V$ and stack them as a 3rd-order representation tensor $\mathcal{Z}\in\mathbb{R}^{I \times I \times V}$. The number of samples $I$ can be further departed by $I_{1}$ and $I_{2}$, where $I= I_{1} \times I_{2}$, $I_{1}$ is the number of samples in each cluster and $I_{2}$ is the number of groups. To better exploit the low rank information, we reshape $\mathcal{Z}$ into a 5th-order tensor with size $(I_{1} \times I_{2} \times I_{3}\times I_{4}\times I_{5} )$, where $I_{1} \times I_{2} = I$, $I_{3} \times I_{4} = I$, $I_{5} = V$, and two of $I_{d}, d = 1 ,\cdots, 4$ are the number of clusters. Furthermore, the tensor network architecture learning framework is applied to exploit the low rank properties and high-order correlations in $\mathcal{Z}$. The final clustering results are obtained by the spectral clustering with the affinity matrix $\mathbf{M}=\frac{1}{V} \sum_{v=1}^{V}\left|\mathbf{Z}_{v}\right|+\left|\mathbf{Z}_{v}^{\mathrm{T}}\right|$.  }
\label{fig:fullflowchart}
\end{figure*}

The overall pipeline is shown in Fig. \ref{fig:fullflowchart}. The main contributions are summarized as follows:
\begin{itemize}
	\item We propose the adaptively topological tensor network (ATTN) framework. It can adaptively search for a suitable topology for a specific instance tensor by pruning redundant connections from an existing tensor network. If there is no or little correlation between two adjacent factors, the edges connecting them will be clipped. 
	\item We formulate the optimization model and develop a resultant algorithm to solve it, which is called the adaptively topological tensor network algorithm (ATTNA). It selects the redundant connections according to their correlations based on the relative standard error appropriately, and the greedy algorithm based rank increment strategy is additionally employed. The image reconstruction experiments demonstrate the effectiveness of our proposed model.
	\item We apply the ATTN to multi-view subspace clustering and develop the adaptively topological tensor network based multi-view subspace clustering (ATTN-MSC). Numerical experiments on six benchmark multi-view datasets demonstrate the superiority of our proposed method in terms of six commonly used metrics in comparison with the state-of-the-art methods.
\end{itemize}

The remainder of this paper is organized as follows. The notations and the related work are briefly introduced in Section~\ref{sec:2}. The optimally topological tensor network framework and its corresponding learning algorithm are detailed in Section~\ref{sec:3}. Section~\ref{sec:4} introduces the ATTN-MVC model with a corresponding optimization algorithm to solve it. The storage complexity and computational complexity of this algorithm are also discussed in Section~\ref{sec:4}. In addition, the experiment results and comprehensive analysis are demonstrated in Section~\ref{sec:5}. Finally, the conclusions are drawn in Section~\ref{sec:6}.  

\section{Notations and Preliminaries}
\label{sec:2}
In this section, we first briefly introduce some notations and preliminaries of tensor operations used in this paper. Then the related work about the tensor based multi-view subspace clustering will be discussed.
\subsection{Notations}
Throughout this paper, we use lower case letters, bold lower case letters, bold upper case letters, and calligraphy letters to denote scalars, vectors, matrices, and tensors, respectively, e.g., $g$, $\mathbf{g}$, $\mathbf{G}$ and $\mathcal{G}$. Frequently used notations are listed in Table~\ref{tab:notations}, where $\mathcal{G}$ is a 3rd-order tensor.

\begin{table}[htbp]
\centering
\caption{Summary of notations in this paper.}
\setlength{\tabcolsep}{3mm}{
	\begin{tabular}{{l|l}}
		\hline
		Denotation         &Definition \\
		\hline
 		$\mathbf{G}_{i}$    		&$\mathbf{G}_{i}=\mathcal{G}(:,:,i)$\\                
		$g_{i_{1},i_{2},i_{3}}$     &The $(i_{1}, i_{2}, i_{3})$-th entry of tensor $\mathcal{G}$\\
		$\mathcal{G}(i_{1},i_{2},i_{3})$     &The $(i_{1}, i_{2}, i_{3})$-th entry of tensor $\mathcal{G}$\\
		$\text{tr}(\mathbf{G})$	&$\text{tr}(\mathbf{G})=\sum_{i=1}^{I}g_{i,i}$	\\
		$\|\mathbf{G}\|_{1}$           &$\|\mathbf{G}\|_{1}=\sum_{i_{1}, i_{2}}\left|g_{i_{1},i_{2}}\right|$ \\    
		$\|\mathbf{G}\|_{\mathrm{F}}$  &$\|\mathbf{G}\|_{\mathrm{F}}=\sqrt{\sum_{i_{1}, i_{2}} g_{i_{1},i_{2}}^{2}}$          \\
		$\|\mathbf{G}\|_{*}$      	&Sum of singular values of $\mathbf{G}$ \\   
		$\|\mathbf{G}\|_{\infty}$      &$\|\mathbf{G}\|_{\infty}=\max _{i_{1}, i_{2}}\left|\mathbf{G}_{i_{1}, i_{2}}\right|$ \\
		$\|\mathbf{G}\|_{2, 1}$       	&$\|\mathbf{G}\|_{2, 1}=\sum_{j}\|\mathbf{G}(:, j)\|_{2}$\\
		$\|\mathcal{G}\|_{\mathrm{F}}$     &$\|\mathcal{G}\|_{\mathrm{F}}=\sqrt{\sum_{i_{1}, i_{2}, i_{3}}\left|g_{i_{1}, i_{2}, i_{3}}\right|^{2}}$   \\	
		\hline	
	\end{tabular}
}
\label{tab:notations}
\end{table}

\subsection{Preliminaries}
\begin{definition}
[\textbf{Mode-$n$ unfolding}]~\cite{kolda2009tensor}
For tensor $\mathcal{G}\in\mathbb{R}^{I_1\times\cdots\times I_N}$, its matricization along the $n$-th mode is denoted as $\mathbf{G}_{(n)} \in \mathbb{R}^{I_{n} \times I_{1} I_{2} \cdots I_{n-1} I_{n+1} \cdots I_{N}}$. 
\end{definition}

\begin{definition}[\textbf{Tensor Contraction}]~\cite{liu2021tensors}
For tensor $\mathcal{A}\in \mathbb{R}^{I_{1}\times \cdots \times I_{N} \times J_{1} \times \cdots \times J_{L}}$ and $\mathcal{B}\in \mathbb{R}^{J_{1}\times \cdots \times J_{L} \times K_{1} \times \cdots \times K_{M}}$, tensor contraction indicates contracting their common indices ${J_{1},\cdots,J_{L}}$ as $\mathcal{C}=\operatorname{TC}(\mathcal{A}, \mathcal{B})\in \mathbb{R}^{I_1 \times \cdots \times I_N \times K_1 \times \cdots \times K_M}$, whose entries are calculated by
\begin{equation}
\begin{aligned}
c_{i_1, \ldots, i_N, k_1, \ldots, k_M}=\sum_{j_{1},\cdots,j_{L}}a_{i_{1},\cdots,i_{N},j_{1},\cdots,j_{L}}b_{j_{1},\cdots,j_{L},k_{1},\cdots,k_{M}}.
\end{aligned}
\end{equation}
 \end{definition}

\subsection{Tensor based multi-view subspace clustering}
Most of the tensor based multi-view subspace clustering models are based on self-representation assumptions, and each sample can be represented as a linear combination of all the samples \cite{liu2022tensor}. Given a mutli-view dataset $\mathbf{X}_{v}\in\mathbb{R}^{C_{v}\times I}~(v=1,\cdots,V)$, where $C_{v}$ is the feature dimensions, $I$ is the number of samples and $V$ is the number of views. Therefore, the general formulation of TMSC models can be written as follows:
\begin{equation}
\begin{aligned}
&\min_{\mathbf{Z}_{v}, \mathbf{E}_{v}} \lambda\mathrm{L}(\mathbf{E})+\mathrm{R}(\mathcal{Z}) \\
&\text{~s. t.~}\mathbf{X}_{v}=\mathbf{X}_{v} \mathbf{Z}_{v}+\mathbf{E}_{v}, v=1, \cdots, V \\
&~~~~~~\mathcal{Z}=\Omega\left(\mathbf{Z}_{1}, \mathbf{Z}_{2}, \cdots, \mathbf{Z}_{V}\right), \\
&~~~~~~~\mathbf{E}=\left[\mathbf{E}_{1} ; \mathbf{E}_{2}; \cdots; \mathbf{E}_{V}\right],
\label{eq:tsvdmsc}
\end{aligned}
\end{equation}
where $\lambda$ denote the trade-off parameter, $\mathrm{L}(\cdot)$ here indicates the loss function and $\mathrm{R}(\cdot)$ is the regularization term, the function $\Omega(\mathbf{Z}_{1}, \mathbf{Z}_{2}, \cdots, \mathbf{Z}_{V})$ constructs the 3rd-order tensor $\mathcal{Z}$ by merging the all self-representation matrices $\mathbf{Z}_{v}, v = 1, \cdots, V $. Various regularization terms are employed in different methods to better exploit the low rank properties of representation tensor $\mathcal{Z}$. 
The Tucker decomposition has been widely employed in TMSC for constraining the self-representation tensor \cite{chen2021low, cheng2018tensor}. In \cite{lu2021multi}, a Kronecker-basis-representation-based tensor sparsity measure based on Tucker decomposition and CP decomposition over $\mathcal{Z}$ is used in TMSC. The hyper-Laplacian regularized Nonconvex Low-rank representation (HNLR) method \cite{wang2022hyper} based on t-SVD-TNN has been proposed to exploit the complementary information among different views. In order to jointly consider the global consensus constraint and local view-specific geometrical regularization for nonlinear subspace clustering, the hyper-Laplacian regularized multi-linear multi-view self-representation (HLR-M$^{2}$VS) \cite{xie2020hyper} is proposed for TMSC. In \cite{sun2020multi}, a tensor log-regularizer based on t-SVD-TNN can better approximate the tensor rank for boosting the clustering performance. In \cite{wang2022hyper}, a non-convex Laplace function replaces the t-SVD-TNN in (\ref{eq:tsvdmsc}) to improve the approximating performance of the global low-rank structure. By additionally applying the hyper-Laplacian regularizer, the hyper-Laplacian regularized Non-convex Low-rank Representation (HNLR) is proposed in \cite{wang2022hyper}. Since the tensor network has a more compact and flexible structure to capture the low rank information \cite{liu2022tensor}, the non-negative tensor ring (NTR) decomposition and the graph regularization non-negative tensor ring (GNTR) decomposition are introduced for TMSC with promising clustering performance. However, the tensor ring only connects the adjacent modes and lacks the capacity of exploring the relationships between non-adjacent modes.

\subsection{Tensor Decomposition}
Tensor decomposition aims to effectively express the high-dimensional data as sparsely interconnected small factor matrices and core tensors. It can be roughly divided into two categories, manual design-based and data dependent model. 

CP decomposition \cite{hitchcock1927expression} is one of the classical manual design-based tensor decomposition models, whose idea is to express a high-order tensor as the sum of a finite number of rank-one factors. Tucker decomposition \cite{tucker1963implications} represents a tensor as a core tensor multiplied by a matrix along each mode. Furthermore, the t-SVD framework \cite{kilmer2008third} can be seen as a different generalization of matrix singular value decomposition. Tensor Train (TT) decomposition, which is also called Matrix product state (MPS) in quantum physics \cite{oseledets2011tensor}, is the simplest and best-known tensor network state. Beyond that, the main idea of TT is to decompose an $N$th-order tensor into $N$ factors, where the head and tail factors are matrices and others are 3rd-order tensors. TR model \cite{zhao2016tensor} replaces the two matrices in the TT factor with third-order tensors and establishes additional operations between them, which is an extended TT form. Since TT and TR decomposition only establishes the operation between two adjacent factors, a new Tensor decomposition, Fully Connected Tensor Network (FCTN) decomposition \cite{zheng2021fully}, is proposed. It decomposes a tensor of $N$th-order into a series of low-dimensional $(N-1)$th-order factors and establishes an operation between any two factors.

As for the data-dependent tensor decomposition model, there are some works have begun to explore the optimal tensor network topology in tensor completion, image denoising, and neural network compression  \cite{ballani2014tree,hashemizadeh2020adaptive,li2020evolutionary,nie2021adaptive}. Ballani et al. propose an agglomerative strategy combined with rank adaptive cross approximation techniques to achieve the tree adaptive approximation in the hierarchical tensor format \cite{ballani2014tree}. A simple greedy approach that gradually increases the tensor rank has been applied in \cite{hashemizadeh2020adaptive} to enhance the performance of tensor decomposition model. Through reforging the graphical representation of tensor network, Li et al. \cite{li2020evolutionary} employ the genetic algorithm to obtain the optimal tensor network structures. The generalized tensor rank is defined in \cite{nie2021adaptive} to characterize the correlations between the two adjacent factors in the tensor network. Based on that, an adaptive tensor network decomposition (ATN) has been proposed.

\section{Adaptively Topological Tensor Network}
\label{sec:3}
In this section, we introduce the generalized tensor network and its relevant concepts. Further, we present the details of our proposed tensor network as well as its corresponding algorithms.
\subsection{Representation}
To accommodate various representation tensors from multi-view data in clustering, we propose the adaptively topological tensor network (ATTN) framework. In this paper, to better determine which edge ranks are important for a tensor network with different possible self-representation tensors, we apply the FCTN architecture. 
Therefore, before introducing our model, we first present the mathematical model of FCTN and several relevant concepts in this section.
 
\begin{definition}[\textbf{Fully Connected Tensor Network (FCTN)}]~\cite{zheng2021fully}
The FCTN represents an $N$th-order tensor by a multi-linear operation on $N$ tensor factors. Given an $N$-th order tensor $\mathcal{X} \in \mathbb{R}^{I_1 \times I_2 \times \cdots \times I_N}$, the FCTN decomposition can be written as

\begin{equation}
\begin{aligned}
&\mathcal{X}\left(i_{1}, i_{2}, \cdots, i_{N}\right)=\operatorname{TC}(\{\mathcal{G}^{(n)}\}_{n=1}^{N})(i_{1}, \cdots, i_{n}, \cdots, i_{N})=\\
&\sum_{r_{1,2}=1}^{R_{1,2}} \sum_{r_{1,3}=1}^{R_{1,3}} \cdots \sum_{r_{1, N}=1}^{R_{1, N}} \sum_{r_{2,3}=1}^{R_{2,3}} \cdots \sum_{r_{2, N}=1}^{R_{2, N}} \cdots \sum_{r_{N-1, N}=1}^{R_{N-1, N}} \\
&\left\{\mathcal{G}^{(1)}\left(i_{1}, r_{1,2}, r_{1,3}, \cdots, r_{1, N}\right) \cdots \right. \\
& ~\mathcal{G}^{(n)}\left(i_{n},r_{1, n}, r_{2, n}, \cdots, r_{n-1, n},  r_{n, n+1}, \cdots, r_{n, N}\right) \cdots \\
&\left.\mathcal{G}^{(N)}\left(i_{N},r_{1, N}, r_{2, N}, \cdots, r_{N-1, N} \right)\right\},
\end{aligned}
\end{equation}
Note that there is only one independent dimension in each core tensor so the index $i_{1}^{n}$ of $n$-th tensor can be abbreviated to $i_{n}$, $1 \leq n \leq N$. Then the edge ranks of FCTN $\mathbf{R}_{\mathrm{FCTN}}$ can be denoted by an $(N-1) \times (N-1)$ matrix as follows:
\begin{equation}
\mathbf{R}_{\mathrm{FCTN}}=\left(\begin{array}{cccc}
R_{1,2} & R_{1,3} & \cdots & R_{1, N} \\
R_{2,1} & R_{2,3} & \cdots & R_{2, N} \\
\vdots & \vdots & \ddots & \vdots \\
R_{N-1,1} & R_{N-1,2} & \cdots & R_{N-1, N}
\end{array}\right),
\end{equation}
where $R_{i, j}$ denotes the dimension of the common modes between $\mathcal{G}^{(i)}$ and $\mathcal{G}^{(j)}$ and satisfies $R_{i, j} = R_{j, i}$ ($1 \leq i<j \leq N, i, j \in \mathbb{N}_{+}$).
\end{definition}

\begin{definition}[\textbf{Redundant Connections}]
\label{def:recon}
Given an $N$th-order tensor represented by a tensor network $\mathcal{X} =\Psi(\mathcal{G}^{(1)},\mathcal{G}^{(2)},\cdots,\mathcal{G}^{(N)})$, the number of connections in this tensor network is $B$, and $N$ is the number of tensor factors. Redundant connections in $\Psi$ are determined by the correlations between the involved mode pairs, which can be reflected by the relative standard error (RSE) defined as follows:
\begin{equation}
\operatorname{RSE}=\frac{\|\mathcal{X}_{\text{new}}-\mathcal{X}\|_{\mathrm{F}}}{\|\mathcal{X}\|_{\mathrm{F}}},
\label{eq:rse1}
\end{equation}
where $\mathcal{X}_{\text{new}}$ is the reconstructed tensor, and $\mathcal{X}$ is the original tensor. The $\operatorname{RSE}^{\Psi}$ denotes the $\operatorname{RSE}$ of the reconstructed tensor by tensor network $\Psi$, and $\operatorname{RSE}^{\Psi-
R_{i,j}} (R_{i,j}= R_{j,i}, 1 \leq i < j \leq N, i, j \in \mathbb{N}_{+})$ means the $\operatorname{RSE}$ of the reconstructed tensor by tensor network $\Psi$ without connection for $R_{i,j}$. The change of these values is calculated by $\operatorname{\delta}^{(b)} = \operatorname{RSE}^{\Psi-
R_{i,j}} - \operatorname{RSE}^{\Psi},~b = 1, \cdots, B$ and $B$ is the number of the edge connection. Unlike the "valuable" connections, whose disappearance will make the $\operatorname{RSE}^{\Psi-
R_{i,j}}$ rise rapidly, the loss of meaningless connections always leads to small $\operatorname{\delta}^{(b)}$. In this way, the redundant connections are  denoted as $\{R_{i,j}\}\in\mathbb{R}^{C}$, where $C$ is the number of redundant connections, whose $\operatorname{\delta}^{(b)}$ values are much smaller than others.
\end{definition}

Given an $N$th-order tensor $\mathcal{X}$, our proposed ATTN will obtain its optimal topology $\Psi(\mathcal{G}^{(1)},\cdots,\mathcal{G}^{(N)})$ with following three features. Firstly, under the premise of that the essential information along each mode can be fully exploited, the $\Psi$ ensures that the number of the tensor factors is minimized. That is, there is no inner tensor factor. In addition, the redundant connections in $\Psi$ are all pruned and the remaining edge rank $\mathbf{R}_{\mathrm{OA}}$ are all the most promising ones.

\subsection{Optimization Model And Algorithm}
Most existing tensor networks are model-driven. However, some recent applications have a huge size of data from different sources. Using the tensor network with pre-defined architecture to approximate these different high-order data tensors may lead to sub-optimal performances and waste data structure information. Therefore, we want to obtain the tensor network architecture of the self-representation tensor that can well fit different data by a data-driven approach, whose requirement can be met by FCTN. Therefore, the optimization problem for ATTN decomposition can be formulated into FCTN decomposition as
\begin{equation}
\min _{ \mathcal{G}^{(1)}, \cdots, \mathcal{G}^{(N)}}\left\|\mathcal{X}-\operatorname{TC}( \{\mathcal{G}^{(n)}\}_{n=1}^{N})\right\|_{\mathrm{F}}.
\label{eq:otn1}
\end{equation}
In order to distinguish it from the acquired adaptive network extension $\Psi$, we denote the initial tensor network structure FCTN as $\tilde{\Psi}$, whose number of the edge connection is $B = N(N-1)/2$. 
The detailed algorithm flow about ATTN (ATTNA) is as follows.

\subsubsection{Initialize the $\tilde{\Psi}$ and update the $\tilde{\Psi}$ with ALS algorithm} We randomly initialize the $\tilde{\Psi}$ with small ranks (but not less than 2 to guarantee full connection), e.g. $ \mathbf{R}_{\tilde{\Psi}}(i,j) = r = 2$, $1 \leq i<j \leq N, i, j \in \mathbb{N}_{+}$. Since the resultant problem in (\ref{eq:otn1}) contains $N$ variables $\{\mathcal{G}^{(n)}\}_{n=1}^{N}$, thus the alternating least squares (ALS) algorithm can be employed to deal with it. The principle of the ALS algorithm is to update each variable in turn while fixing other variables until all variables meet the convergence criterion~\cite{kolda2009tensor}. Therefore, the closed solution of each sub-problem is
\begin{equation}
\mathbf{G}_{(1)}^{(n)}=\argmin _{\mathbf{G}_{(1)}^{(n)}} \left\|\mathbf{X}_{(n)}-(\mathbf{G}_{(1)}^{(n)})(\mathbf{A}^{\neq n}_{(n)})\right\|_{\mathrm{F}},
\label{eq:als}
\end{equation}
where $\mathcal{A}^{\neq n} =\operatorname{TC}(\mathcal{G}^{(1)}, \cdots,\mathcal{G}^{(n-1)}, \mathcal{G}^{(n+1)},\cdots,\mathcal{G}^{(N)})$, and $\mathbf{A}^{\neq n}_{(n)}$ is the mode-$n$ unfolding of $\mathcal{A}^{\neq n}$.

\subsubsection{Cut redundant connections} The process of cutting redundant connections can be further put into two parts. Firstly, select the redundant connections by referring to Definition \ref{def:recon}. The $\operatorname{RSE}^{\tilde{\Psi}}$ and $\operatorname{RSE}^{\tilde{\Psi}-R_{i,j}}$ $(1 \leq i < j \leq N, i, j \in \mathbb{N}_{+})$ represent the relative standard error of reconstructed tensor from $N$th-order network $\tilde{\Psi}$, and the $\tilde{\Psi}$ without the edge of $R_{i,j} \in \mathbf{R}_{\tilde{\Psi}}$ (i.e., set $R_{i,j} = 1$), respectively. Secondly, the set of redundant connections $\{R_{i,j}\}~(1 \leq i < j \leq N, i, j \in \mathbb{N}_{+})$ with size $C$, can be acquired by calculating the $\operatorname{{\delta}^{(b)}} = \operatorname{RSE}^{\tilde{\Psi}-R_{i,j}} - \operatorname{RSE}^{\tilde{\Psi}}$, $b = 1, \cdots, B$ in turn, where $B$ is the number of the edge connection. Apart from that, the values of those chosen ranks $\{R_{i,j}\}$ will be set to one. 

\subsubsection{Rank increment strategy with greedy algorithm}
A critical problem of equation (\ref{eq:otn1}) is to determine the edge ranks automatically. In this work, we adopt a logical heuristic algorithm, namely greedy search \cite{barron2008approximation, shalev2011large}, to choose the most promising ranks. The main idea of greedy algorithms is to solve the optimization problem by making the locally optimal choice at each stage with the hope of leading to a globally optimum solution \cite{cormen2022introduction}. For an $N$th-order tensor network with $C$ redundant connections, there are $B-C$ choices for each iteration to increase the rank gradually.

The edge ranks without the redundant ones are put into a matrix denoted as $\mathbf{R}_{\mathrm{OA}}\in\mathbb{R}^{(N-1)\times(N-1)}$. At each iteration, the $R_{i,j} \in \mathbf{R}_{\mathrm{OA}}$ $(1 \leq i < j \leq N, i, j \in \mathbb{N}_{+})$ increases by step size $a$ in turn. Thus, $R_{i,j}=R_{i,j}+a$, and the involved cores $\mathcal{G}^{(i)}$ and $\mathcal{G}^{(j)}$ will add $a$ new slices correspondingly, represented as $\tilde{\mathcal{G}}^{(i)}$ and $\tilde{\mathcal{G}}^{(j)}$. Additionally taking few iterations to optimize the $\operatorname{RSE}^{(b_{1})}$ for each possible edge rank $R_{i,j}$ by 
\begin{equation}
 \frac{\left\|\mathrm{TC}(\{\mathcal{G}^{(1)},\cdots,\tilde{\mathcal{G}}^{(j)},\cdots,\mathcal{G}^{(N)}\})-\mathrm{TC}(\{\mathcal{G}^{(n)}\}_{n=1}^{N})\right\|_{\mathrm{F}}}{{\left\|\mathrm{TC}(\{\mathcal{G}^{(n)}\}_{n=1}^{N})\right\|_{\mathrm{F}}}},   
\end{equation}
where $b_{1} = 1, \cdots,(B-C)$. Based on that, the edge rank $R_{i,j}$ which leads to the steepest decrease referring to $\operatorname{RSE}^{(b_{1})}$ is selected to increase finally.  

The ALS estimation and the rank incremental steps are repeated
until the error, i.e., $\left\|\mathcal{X}_{\text{new}}-\mathcal{X}\right\|_{\mathrm{F}}/{\left\|\mathcal{X}\right\|_{\mathrm{F}}}$, becomes smaller than a pre-defined threshold $\varepsilon$, or the number of iterations reaches its maximum. 

Note that, the rank-1 connection means that it has been removed before the tensor network contraction since such an edge is unnecessary and will not affect the computation consistency. Hence, two connected factors indicate the corresponding edge rank is larger than one, and vice versa.

Due to the limitations of the space, the details about the optimization procedure of these strategies has illustrated in our supplementary material. Since our proposed method prunes the redundant connections and enhances the new tensor network representation capacity simultaneously, the ATTNA can automatically search the optimal architecture for different input tensors.

\subsection{Complexity Analysis}
Because the ATTNA heavily relies on the ALS algorithm, the computational complexity  mainly comes from two parts, i.e., the tensor contractions of $\mathcal{A}^{\neq n}$ and the matrix inversion for calculating $\mathbf{G}_{(1)}^{(n)}$. The former one includes $(N-2)$ matrix multiplication and conduct the complexity is $\operatorname{O}(Nr^{B}I^{N-1})$. Apart from that, the computational cost of the matrix inversion is $\operatorname{O}(Nr^{B}I^{N})$. While the number of edge ranks in ATTNA is $B$ and their values are all pre-defined as the same and small value $r$, the number of edge ranks in the greedy algorithm based rank-incremental strategy equals $(B-C)$ and its rank is denoted as $R_{\mathrm{OA}}$. Therefore, according to the aforementioned discussion, its complexity is $\operatorname{O}(NR_{\mathrm{OA}}^{B-C}I^{N})$. Regardless of the iteration number, the whole computational complexity of ATTNA is $\operatorname{O}(NI^{N}(R_{\mathrm{OA}}^{B-C}+r^{B}))$. 

On the other hand, the storage cost of ATTNA is composed of the size of $\{\mathcal{G}^{(n)}\}_{n = 1}^{N}$. Denote that the average number of connections that each factor $\mathcal{G}^{(n)}$ contains is $c$, thus the whole storage cost is $\operatorname{O}(NIR_{\mathrm{OA}}^{c})$.

\section{The Adaptively Topological Tensor Network For Multi-view Subspace Clustering}
\label{sec:4}
\subsection{Problem Formulation}
Given the multi-view dataset $\mathbf{X}_{v}\in \mathbb{R}^{C_{v} \times I}$, $v=1,\cdots,V$, where $C_{v}$ is the feature dimension, $V$ and $I$ are the numbers of views and samples, respectively. Applying the optimally topological tensor network based constraints, the self-representation based multi-view subspace clustering model (MSC) can be reformulated as
\begin{equation}
\begin{aligned}
&\min _{\mathbf{Z}_{v}, \mathbf{E}_{v}, \{\mathcal{G}^{(n)}\}_{n=1}^{N}} \lambda\|\mathbf{E}\|_{2,1}+\|\mathcal{Z}-\Phi(\Psi(\{\mathcal{G}^{(n)}\}_{n=1}^{N}),[I,I,V])\|_{\mathrm{F}}^{2} \\
&~~~~~~~~\text {s. t.~~} \mathbf{X}_{v}=\mathbf{X}_{v} \mathbf{Z}_{v}+ \mathbf{E}_{v}, v=1, \cdots, V, \\
&~~~~~~~~~~~~~~~\mathcal{Z}=\Omega\left(\mathbf{Z}_{1}, \mathbf{Z}_{2}, \cdots, \mathbf{Z}_{V}\right),\\
&~~~~~~~~~~~~~~~\mathbf{E}=\left[\mathbf{E}_{1} ; \mathbf{E}_{2} ; \cdots; \mathbf{E}_{V}\right],
\end{aligned}
\label{eq:otnmsc}
\end{equation}
where function $\Omega(\mathbf{Z}_{1}, \mathbf{Z}_{2}, \cdots, \mathbf{Z}_{V})$ constructs the 3rd-order tensor $\mathcal{Z}$ by merging all self-representation matrices $\mathbf{Z}_{v},v=1,\cdots,V$. Furthermore, function $\Phi(\mathcal{T},\mathbf{s})$ reconstructs the tensor $\mathcal{T}$ with the size vector $\mathbf{s}$, and $N$ is the number of tensor factors. $\Psi(\{\mathcal{G}^{(n)}\}_{n=1}^{N})$ is denoted as our proposed optimal tensor network.

Since the total number of samples in MSC is $ I = I_{1} \times I_{2}$, where $I_{1}$ is the clusters amount and $I_{2}$  equals the number of samples in each cluster. Considering that the performance of exploiting the low rank information in higher-order tensor is more attractive, we reorganize the self-representation tensor $\mathcal{Z}\in \mathbb{R}^{I \times I \times V}$ as a 5th-order tensor with size $I_{1} \times I_{2} \times I_{3} \times I_{4} \times V$, where $I_{1} \times I_{2} = I$, $I_{3} \times I_{4} = I$, and two of $I_{d}$ ($d = 1,\cdots,4$) are the numbers of clusters. Meanwhile, $N = 5$.

\subsection{Solution}
The above optimization problem can be solved by leveraging the augmented Lagrange multiplier (ALM) \cite{lin2010augmented}. Additionally employing an extra variable $\mathcal{S}$, the optimization problem in (\ref{eq:otnmsc}) can be rewritten as the minimization of the following term:
\begin{equation}
\begin{aligned}
&\mathrm{L}\left(\mathbf{Z}_{1}, \cdots, \mathbf{Z}_{V} ; \mathbf{E}_{1}, \cdots, \mathbf{E}_{V} ; \mathcal{S}\right) \\
&=\lambda\|\mathbf{E}\|_{2,1}+\|\mathcal{S}-\Phi\left(\Psi(\{\mathcal{G}^{(n)}\}_{n=1}^{N}),[I,I,V]\right)\|_{\mathrm{F}}^{2}\\
&+\sum_{v=1}^{V}(\left\langle\mathbf{Y}_{v}, \mathbf{X}_{v}-\mathbf{X}_{v}\mathbf{Z}_{v}-\mathbf{E}_{v}\right\rangle+\frac{\mu}{2}\|\mathbf{X}_{v}-\mathbf{X}_{v} \mathbf{Z}_{v}-\mathbf{E}_{v}\|_{\mathrm{F}}^{2})\\
&+\langle\mathcal{W}, \mathcal{Z}-\mathcal{S}\rangle+\frac{\rho}{2}\|\mathcal{Z}-\mathcal{S}\|_{\mathrm{F}}^{2}\\
&~~~~~\text {s. t.~~} \mathcal{Z}=\Omega\left(\mathbf{Z}_{1}, \mathbf{Z}_{2}, \cdots, \mathbf{Z}_{V}\right).
\end{aligned}
\label{eq:alm}
\end{equation}

Here $\mu$ and $\rho$ are penalty parameters. $\mathbf{Y}_{v}$ and $\mathcal{W}$ are Lagrange multipliers. We alternately update each variable while fixing the other variables to optimize the above term. Hence, the problem in (\ref{eq:alm}) can be turned into solving the following four sub-problems alternately, and their corresponding solution is discussed in detail as follows.

\subsubsection{Sub-problem with respect to $\mathcal{Z}$} We keep the values of the other parameters $\mathbf{E}$, $\mathcal{S}$, $\mathbf{Y}_{v}$ and $\mathcal{W}$, the sub-problem for $\mathcal{Z}$ is
\begin{equation}
\begin{aligned}
\min _{\mathbf{Z}_{v}}&\left\langle\mathbf{Y}_{v}, \mathbf{X}_{v}-\mathbf{X}_{v} \mathbf{Z}_{v}-\mathbf{E}_{v}\right\rangle+\frac{\mu}{2} \| \mathbf{X}_{v}-\mathbf{X}_{v} \mathbf{Z}_{v}-\mathbf{E}_{v}\|_{\mathrm{F}}^{2}\\
&+\left\langle\mathbf{W}_{v},\mathbf{Z}_{v}-\mathbf{S}_{v}\right\rangle+\frac{\rho}{2}\| \mathbf{Z}_{v}-\mathbf{S}_{v} \|_{\mathrm{F}}^{2}.
\end{aligned}
\label{eq:subz}
\end{equation}
By setting the derivative of the objective function in (\ref{eq:subz}) with respect to $\mathbf{Z}_{v}$ to zero, we can obtain 
\begin{equation}
\begin{aligned}
\mathbf{Z}_{v}^{t+1}=&(\mathbf{I}+\frac{\mu}{\rho} \mathbf{X}_{v}^{\mathrm{T}} \mathbf{X}_{v})^{-1}((\mathbf{X}_{v}^{\mathrm{T}} \mathbf{Y}_{v}+\mu \mathbf{X}_{v}^{\mathrm{T}} \mathbf{X}_{v}-\\
&\mu \mathbf{X}_{v}^{\mathrm{T}} \mathbf{E}_{v}-\mathbf{W}_{v}) /\rho+\mathbf{S}_{v}).\\
\label{eq:otnmsc1}
\end{aligned}
\end{equation}

\subsubsection{Sub-problem with respect to $\mathbf{E}_{v}$} With fixed other variables, we solve the following problem for the purpose of updating $\mathbf{E}_{v}, v = 1, \cdots, V$:
\begin{equation}
\begin{aligned}
\mathbf{E}^{t+1}=& \argmin _{\mathbf{E}} \lambda\|\mathbf{E}\|_{2,1}+\sum_{v=1}^{V}\bigg(\left\langle\mathbf{Y}_{v}^{t}, \mathbf{X}_{v}-\mathbf{X}_{v} \mathbf{Z}_{v}^{t+1}-\mathbf{E}_{v}\right\rangle \\
&+\frac{\mu}{2}\left\|\mathbf{X}_{v}-\mathbf{X}_{v} \mathbf{Z}_{v}^{t+1}-\mathbf{E}_{v}\right\|_{\mathrm{F}}^{2}\bigg) \\
=&\argmin _{\mathbf{E}} \frac{\lambda}{\mu}\|\mathbf{E}\|_{2,1}+\frac{1}{2}\|\mathbf{E}-\mathbf{D}\|_{\mathrm{F}}^{2},
\end{aligned}
\end{equation}
where $\mathbf{D}$ is constructed by vertically concatenating the matrices $\mathbf{X}_{v}-\mathbf{X}_{v} \mathbf{Z}_{v}^{t+1}+(1 / \mu) \mathbf{Y}_{v}^{t} (v = 1,\cdots,V)$ along column. According to \cite{tang2018learning}, the above problem has the following closed solution:
\begin{equation}
\mathbf{E}_{:, i}^{t+1}=\left\{\begin{array}{l}
\frac{\left\|\mathbf{D}_{:, i}\right\|_{2}-\frac{\lambda}{\mu}}{\left\|\mathbf{D}_{:, i}\right\|_{2}} \mathbf{D}_{:, i}, \quad\left\|\mathbf{D}_{:, i}\right\|_{2}>\frac{\lambda}{\mu} \\
\mathbf{0} \text { otherwise. }
\end{array}\right.
\label{eq:otnmsc2}
\end{equation}

\subsubsection{Sub-problem with respect to $\mathcal{S}$} When $\mathbf{E}$, $\mathcal{Z}$, $\mathbf{Y}_{v}$ and $\mathcal{W}$ are fixed, $\mathcal{S}^{t+1}$ can be obtained by solving the following minimization problem:
\begin{equation}
 \begin{aligned}
\mathcal{S}^{t+1}=\argmin _{\mathcal{S}} &\left\|\mathcal{S}-\Phi\left(\Psi(\{\mathcal{G}^{(n)}\}_{n=1}^{N}),[I,I,V]\right)\right\|_{\mathrm{F}}^{2}\\
&+\frac{\rho}{2}\left\|\mathcal{S}-\left(\mathcal{Z}^{t+1}+\frac{1}{\rho} \mathcal{W}^{t}\right)\right\|_{\mathrm{F}}^{2}.\\ 
 \end{aligned}
\end{equation}
It can be clearly seen that the minimum of the above problem can approach 0 infinitely when $\mathcal{S} = \Phi\left(\Psi(\{\mathcal{G}^{(n)}\}_{n=1}^{N}),[I, I, V]\right) \approx \mathcal{Z}^{t+1}+\frac{1}{\rho} \mathcal{W}^{t}$. Based on that, we transfer it into the following formulation:
\begin{equation}
\begin{aligned}
&\min _{\mathcal{S}}\frac{\rho}{2}\left\|\mathcal{S}-\left(\mathcal{Z}^{t+1}+\frac{1}{\rho} \mathcal{W}^{t}\right)\right\|_{\mathrm{F}}^{2} \\
&~\text {s. t. } \mathcal{S}=\Phi\left(\Psi(\{\mathcal{G}^{(n)}\}_{n=1}^{5}),[I,I,V]\right).\\
\end{aligned}
\label{eq:newpro}
\end{equation}

On this basis, the problem in (\ref{eq:newpro}) can be easily solved by additionally utilizing ATTNA. Through reconstructing the tensor $(\mathcal{Z}+\frac{1}{\rho} \mathcal{W})$ into a 5th-order tensor $\mathcal{F}$ with size $I_{1} \times I_{2} \times I_{3}\times I_{4}\times V$, the core tensors $\tilde{\mathcal{G}}^{(1)}, \cdots,\tilde{\mathcal{G}}^{(5)}$ can be obtained by employing the $\mathcal{F}$ into ATTNA. Afterwards, the low rank representation tensor $\mathcal{S}^{t+1}$ is updated by 
\begin{equation}
\mathcal{S}^{t+1}=\Phi(\operatorname{TC}(\{\tilde{\mathcal{G}}^{(n)}\}_{n=1}^{5}),[I,I,V]).
\label{eq:otnmsc3}
\end{equation}

Note that the ATTN $\Psi$ of $\mathcal{F}$ will be learned at intervals of t at the $t$-th iteration, e.g., interval $t=5$. In subsequent iterations, we apply the learned $\Psi$ to approximate the tensor $\mathcal{F}$ by the 4th-17th steps in ATTNA.

\subsubsection{Update Lagrangian multipliers} After renewing the other parameters, the Lagrangian multipliers can be updated by 
\begin{equation}
\label{eq:otnmsc5}
\begin{split}
\mathbf{Y}_{v}^{t+1}&=\mathbf{Y}_{v}^{t}+\mu\left(\mathbf{X}_{v}-\mathbf{X}_{v} \mathbf{Z}_{v}^{t+1}-\mathbf{E}_{v}^{t+1}\right).\\
\mathcal{W}^{t+1}&=\mathcal{W}^{t}+\rho(\mathcal{Z}^{t+1}-\mathcal{S}^{t+1}).
\end{split}
\end{equation}

Repeatedly updating the aforementioned variables, the representation matrices $\mathbf{Z}_{v}$ $(v = 1, \cdots, V)$ can be obtained accordingly. Furthermore, we utilize  $\mathbf{M}=\frac{1}{V} \sum_{v=1}^{V}\left|\mathbf{Z}_{v}\right|+\left|\mathbf{Z}_{v}^{\mathrm{T}}\right|$ as the affinity matrix, which will be fed into the spectral clustering algorithms for the final clustering results.

The whole optimization procedure of the adaptively topological tensor network for the multi-view subspace clustering (ATTN-MSC) method is briefly summarized in Algorithm \ref{alg:algootnmsc}. In addition, the graphical illustration of the pipeline about the ATTN-MSC has been shown in Fig. \ref{fig:fullflowchart} for better understanding.

\begin{algorithm}[H]
	\caption{ATTN for Multi-view Subspace Clustering}
		\begin{algorithmic}[1]
			\STATE\textbf{Input:} Multi-view data $\{\mathbf{X}_{v}\},v=1,\cdots,V$, tensor size $[I_1,I_2,I_3,I_4,V]$, $\varepsilon$, $B=10$;
			\STATE\textbf{Initialize:} $\mathcal{S}=\mathcal{W}=0, \mathbf{Z}_{v}=0, \mathbf{E}_{v}=0$; $\rho=10^{-4}$, $\mu=10^{-5}$, $\text{tol}=10^{-7}$, $\eta = 2$, $t=0$, $\mu_{\max }=\rho_{\max }=10^{10}$;
			\STATE \textbf{while} $t \leq \text{iter}_\text{max}$ \textbf{do}
			\STATE \textbf{for} $v=1$ to $V$ \textbf{do}
			\STATE \quad Update $\mathbf{Z}_{v}^{t+1}$ via equation (\ref{eq:otnmsc1});
			\STATE \textbf{end for}
			\STATE Update $\mathbf{E}^{t+1}$ by equation (\ref{eq:otnmsc2});
			\STATE \textbf{for} $v=1$ to $V$ \textbf{do}
			\STATE \quad Update $\mathbf{Y}_{v}^{t+1}$ via equation (\ref{eq:otnmsc5});
			\STATE \textbf{end for}		
			\STATE Obtain $\mathcal{Z}^{t+1}=\Phi(\Omega\left(\mathbf{Z}_{1}^{t+1}, \cdots, \mathbf{Z}_{V}^{t+1}\right), [I_{1},I_{2},I_{3},I_{4},V])$;
			\STATE Update $\mathcal{S}^{t+1}$ through ATTNA;
			\STATE Obtain $\mathcal{Z}^{t+1}=\Phi(\mathcal{Z}^{t+1},[I,I,V])$;
			\STATE Update $\mathcal{W}^{t+1}$ by equation (\ref{eq:otnmsc5});
			\STATE $\mu=\min \left(\eta \mu, \mu_{\max }\right)$, $\rho=\min \left(\eta \rho, \rho_{\max }\right)$;
			\STATE \textbf{if} $\max(\left\|\mathbf{X}_{v}-\mathbf{X}_{v} \mathbf{Z}_{v}^{t+1}-\mathbf{E}_{v}^{t+1}\right\|_{\infty} ,\|\mathcal{Z}^{t+1}-\mathcal{S}^{t+1}\|_{\infty}) \leq \text{tol}$
			\STATE \quad break;
			\STATE \textbf{end if} 
			\STATE $t=t+1$;
			\STATE \textbf{end while}
			\STATE Calculate affinity matrix: $\mathbf{M}=\frac{1}{V} \sum_{v=1}^{V}\left|\mathbf{Z}_{v}\right|+\left|\mathbf{Z}_{v}^{\mathrm{T}}\right|$;
			\STATE Apply the spectral clustering method with the affinity matrix $\mathbf{M}$;
			\STATE \textbf{Output:} Clustering result.
	\end{algorithmic}	
\label{alg:algootnmsc}
\end{algorithm}

\subsection{Computational Complexity Analysis}
The whole complexity of the ATTN-MSC algorithm mainly comes from three parts. The cost of calculating the self-representation tensor $\mathcal{Z}$ is $\operatorname{O}\left(V I^{3}\right)$. As for solving the sub-problem concerning the $\mathbf{E}$, it contains the matrix inverse and the matrix multiplication with cost $\mathcal{O}\left(V I^{2}\right)$. The computational complexity of updating the $\mathcal{S}$ is $\operatorname{O}\left(VI^{2} (r^{10}+R^{10-C})\right)$ for ATTN algorithm.

Therefore, the total complexity of TNAL-MSC is $\operatorname{O}\left(TV I^{3}+VI^{2}r^{10}+TVI^{2}R^{10-C}+TVI^{2}\right)$, where $T$ is the number of iterations for ATTN-MSC, $r$ is the pre-defined edge rank, and $R$ is the rank of $\mathcal{S}$.

\section{Experiments and Analysis}
\label{sec:5}
In this section, we process the image reconstruction experiments to indicate the ATTNA's advantages over exploring the low-rank properties. Furthermore, in order to demonstrate the superiority of our proposed model ATTN-MSC, we compare it with two single-view methods, seven matrix-based multi-view models, and four tensor-based multi-view methods on six different applications. 

\subsection{Low-Rank Analysis}
Five real-world images\footnote{http://sipi.usc.edu/database/database.php?volume=misc} have been utilized to demonstrate the superiority of ATTNA by conducting image reconstruction experiments. They are renamed as "Lena", "Peppers", "House", "Sailboat", and "Baboon" in this paper, whose sizes are all equal to $256 \times 256 \times 3$. And the graphical illustration is shown in Fig. 1 in the supplementary material. Since it will be utilized for 5th-order tensors in multi-view subspace clustering further, the original forms of them have been transformed into $16 \times 16 \times 16 \times 16 \times 3$ tensors in this section.

In this section, the effectiveness and low-rank properties of the algorithms for image reconstruction have been compared with each other. The corresponding algorithms about Tucker decomposition, TT-ALS, TR-ALS, FCTN-ALS, ATN and ATTNA will be all applied to present the image tensors, under almost the same RSE. Furthermore, the storage costs of the different tensor decompositions have been calculated and presented in Table~\ref{tab:reconre}. In addition to the time needed for pre-processing the images, and calculating the storage costs, the running time concluded in Table~\ref{tab:reconre} only contains the main body in decomposition and reconstruction. Considering that the speed of the iterative algorithm primarily affects the running time of those methods, we set the maximum number of iterations for all methods except ATN to 300 in our experiment. Besides, the real iteration number of those methods has been concluded in Table \ref{tab:iter}.

As the results in Table~\ref{tab:reconre} show, under the same conditions, the running time of ATTNA is almost at the second-best, but the performance about storage cost has achieved the best one. It can be seen from this that our proposed model will better exploit the low rank information compared to the other decomposition. 


\begin{table}[htbp]
\begin{center}
\caption{Comparison results about Tucker, TT, TR, FCTN, ATN, and ATTN algorithms based image reconstruction experiments on five real word images.}
\label{tab:reconre}
\scalebox{1.1}{
\begin{tabular}{{ccccc}}
\toprule		
Dataset &Method &RSE &Time(s) &Storage Cost 
\\
\midrule			
	   	&Tucker     &0.137      &35.37	            &30649\\
	   	&TT         &0.137      &\textbf{2.89}      &4047\\
Lena	&TR 	    &0.136      &119.20	            &\underline{1668}\\
        &FCTN	    &0.137	    &14.43	            &2064\\
        &ATN        &0.137      &23.17              &2438\\
        &ATTN	    &0.135	    &\underline{9.91}	&\textbf{1356}\\
\midrule			
	   	&Tucker     &0.148 	    &74.33	            &21785\\
	   	&TT         &0.150      &\textbf{10.84}     &7439\\
Peppers	&TR 	    &0.150 	    &310.31	            &3352\\
        &FCTN	    &0.149	    &28.40	            &3852\\
        &ATN        &0.147      &49.49              &\underline{3116}\\
        &ATTN	    &0.150	    &\underline{14.91}	&\textbf{2610}\\
\midrule			
	   	&Tucker     &0.103 	    &28.83	            &16739\\
	   	&TT         &0.098      &\textbf{6.86}      &6095\\
House	&TR 	    &0.098 	    &223.72	            &\underline{2591}\\
        &FCTN	    &0.097	    &25.44	            &3228\\
        &ATN        &0.100      &51.31              &3116\\
        &ATTN	    &0.098	    &\underline{20.66}	&\textbf{2130}\\
\midrule			
	   	    &Tucker     &0.138 	    &56.16	            &32741\\
	   	    &TT         &0.138      &\underline{20.03}  &9295\\
Sailboat	&TR 	    &0.138 	    &\textbf{10.23}	    &8766\\
            &FCTN	    &0.138	    &31.23	            &\underline{4428}\\
            &ATN        &0.137      &54.97              &5190\\
            &ATTN	    &0.138	    &24.06	            &\textbf{3708}\\
\midrule			
	   	&Tucker     &0.161 	    &47.49	            &21721\\
	   	&TT         &0.161      &\textbf{14.66}     &7439\\
Baboon	&TR 	    &0.160 	    &302.93	            &\underline{3528}\\
        &FCTN	    &0.160	    &33.42	            &4194\\
        &ATN        &0.161      &38.95              &7302\\
        &ATTN	    &0.161	    &\underline{20.17}	&\textbf{3058}\\
\bottomrule
\end{tabular}
}
\end{center}
\end{table}

\begin{table}[htbp]
\begin{center}
\caption{The real iteration number of TT, TR, FCTN, ATN and ATTN algorithms in image reconstruction experiments, respectively.}
\label{tab:iter}
\scalebox{1.2}{
\begin{tabular}{{cccccc}}
\toprule		
Methods     &TT     &TR     &FCTN   &ATN    &ATTN \\
\midrule			
Iterations  &2	    &2/300  &300	&5000   &300\\
\bottomrule
\end{tabular}
}
\end{center}
\end{table}

\subsection{Multi-view datasets And Compared Clustering Algorithms}
We evaluate the effectiveness of ATTN-MSC on six multi-view datasets, which cover six different applications, including text clustering, generic object clustering, facial clustering, tree image clustering and scene clustering. The statistics information is summarized in Table \ref{tab:data}, for details please refer to supplementary materials. Note that the number of the seven categories in Caltech101-7 is seriously unbalanced, which ranges from 34 to 798, and the sample size is 1474, which is not divisible by the number of categories, bringing the challenge to reshape the 3rd-order self-representation tensor into a 5th-order tensor. To handle these issues, we choose the size of $22\times67\times22\times67\times6$ to balance tensor reshaping and multi-view information mining.
\begin{table*}[htbp]
\centering
\caption{Statistics of different multi-view datasets.}
\scalebox{0.8}{
    	\begin{tabular}{{cccccc}}
    		\toprule
    		datasets  	&Samples(I) 	&Views(V)  	&Clusters 	&Objective 	&5-D ($I_{1},\cdots,I_{4},V$)\\
    		\midrule
    		Yale		&165  	&3   	&15 &Face 		&(11,15,11,15,3)\\
            UCI Digits    &2000   &3  &10 &Digit  &(200,10,200,10,3)\\
    		Reuters    &1200    &5      &6  &Text       &(200,6,200,6,5)\\
    		100leaves  &1600 &3 &100 &Tree &(100,16,100,16,3)\\
    		Caltech101-7 &1474 &6 &7 &Object &(22,67,22,67,6)\\
    		MSRCV1 &210  &6 &7  &Scene &(30,7,30,7,6)\\
    		\bottomrule	
    	\end{tabular}
}
\label{tab:data}
\vspace{-0.3cm}
\end{table*}

\begin{table*}[htbp]
\begin{center}
\caption{Clustering results on Yale dataset. For ATTN-MSC, we set $\lambda=0.5$.}
\label{tab:yale}
\scalebox{0.6}{
\begin{tabular}{{cccccccccc}}
\toprule		
&Method &F-score &Precision &Recall &NMI &AR &ACC
\\
	\midrule
		&\text{SPC}$_{best}$ 		& 0.138(0.003) 		&0.078(0.002) 	&0.571(0.011)	&0.319(0.015)	&0.034(0.004)	&0.216(0.008)\\			
		&\text{LRR}$_{best}$ 		& 0.554(0.025)		&0.510(0.030)	&0.606(0.019)	&0.741(0.014)	&0.522(0.027)	&0.731(0.022)\\
	\midrule
		&Co-reg 		&0.520(0.066)		&0.487(0.073)	&0.558(0.057)	&0.699(0.043)	&0.486(0.071)	&0.642(0.066)\\
		&DiMSC		&0.564(0.002)	&0.543(0.001)	&0.586(0.003)	&0.727(0.010)	&0.535(0.001)	&0.709(0.003)\\
		&RMSC 		&0.510(0.086)		&0.489(0.092)	&0.535(0.079)	&0.686(0.060)	&0.477(0.092)	&0.638(0.097) \\
		&LMSC 		&0.519(0.003)		&0.475(0.006)	&0.572(0.000)	&0.717(0.002)	&0.484(0.004)	&0.679(0.007)\\
		&ECMSC	&0.460(0.008)	&0.407(0.010)	&0.530(0.014)	&0.685(0.006)	&0.420(0.009)	&0.679(0.028)\\	
 		&LTMSC 		&0.620(0.009)		&0.599(0.011)	&0.643(0.006)	&0.764(0.006)	&0.594(0.009)	&0.736(0.004)\\
		&t-SVD-MSC 	&\underline{0.902(0.066)}		&\underline{0.891(0.075)}	&\underline{0.915(0.058)}	&\underline{0.946(0.036)}	&\underline{0.896(0.071)}	&\underline{0.934(0.053)}\\
		&ETLMSC	&0.542(0.055)	&0.509(0.053)	&0.580(0.061)	&0.706(0.043)	&0.510(0.059)	&0.654(0.044)\\
		&HLR-M$^{2}$VS	&0.695(0.010)	&0.673(0.011)	&0.718(0.010)	&0.817(0.006)	&0.674(0.011)	&0.772(0.012)\\
		&CoMSC &0.627(0.000)	&0.607(0.000)	&0.953(0.000)	&0.762(0.000)	&0.602(0.000)	&0.739(0.000)\\
		&JSMC &0.631(0.000)	&0.581(0.000)	&0.691(0.000)	&0.787(0.000)	&0.605(0.000)	&0.752(0.000)\\
		&ATTN-MSC	&\textbf{1.000(0.000)} &\textbf{1.000(0.000)} &\textbf{1.000(0.000)} &\textbf{1.000(0.000)} &\textbf{1.000(0.000)} &\textbf{1.000(0.000)}\\
\bottomrule
\end{tabular}
 }
\end{center}
\end{table*}

\begin{figure*}[htbp]
\captionsetup[subfigure]{aboveskip=-1pt,belowskip=-1pt}
\centering
\subfloat[ATTN-MSC]{
\begin{minipage}[t]{0.5\linewidth}
\centering
\includegraphics[scale=0.45]{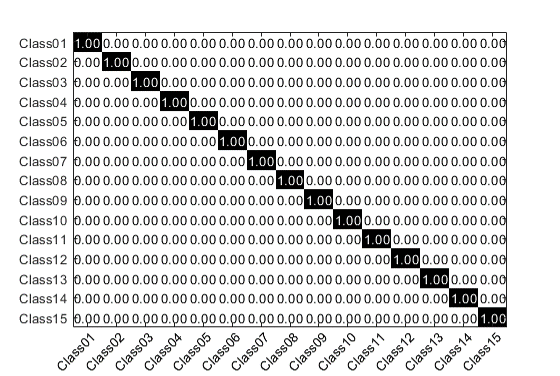}
\end{minipage}%
}%
\subfloat[t-SVD-MSC]{
\begin{minipage}[t]{0.5\linewidth}
\centering
\includegraphics[scale=0.45]{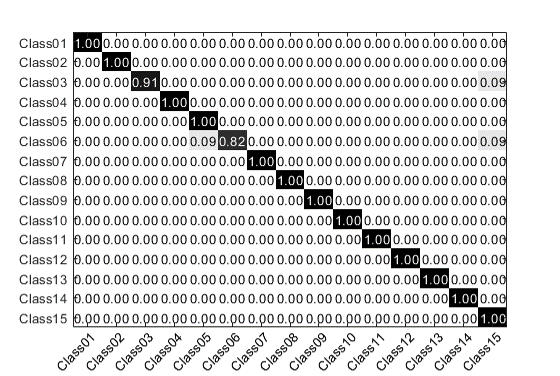}
\end{minipage}%
}%
\centering
\caption{The graphical illustration of the confusion matrices about the Yale datasets over the ATTN-MSC and the t-SVD-MSC which have achieved the top two clustering performances.}
\label{fig:conmatrix}
\end{figure*}

\begin{figure*}[htbp]
\centering
\subfloat[LTMSC]{
\begin{minipage}[t]{0.3\linewidth}
\centering
\includegraphics[scale=0.27]{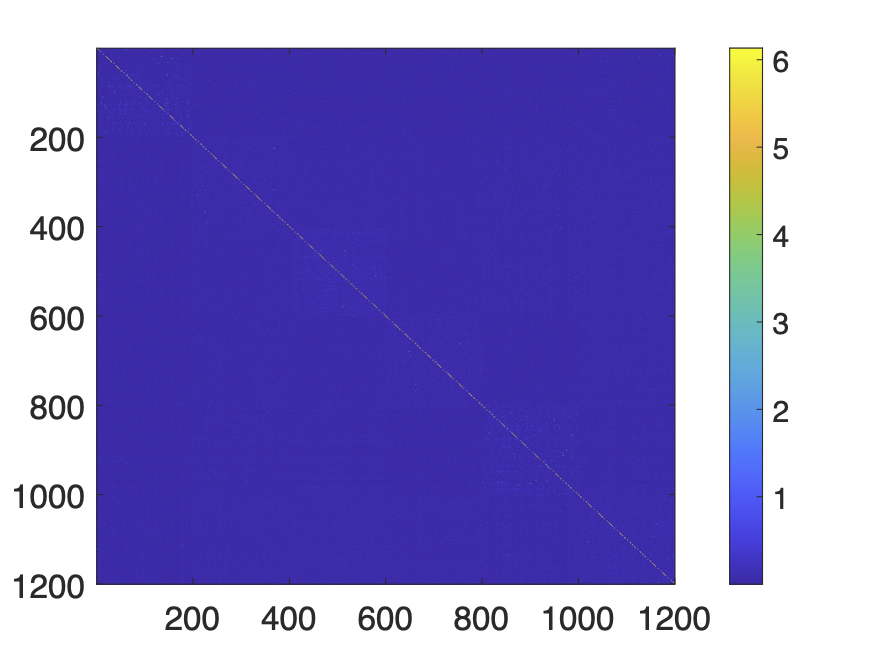}
\end{minipage}%
}%
\subfloat[HLR-M$^{2}$SC]{
\begin{minipage}[t]{0.3\linewidth}
\centering
\includegraphics[scale=0.27]{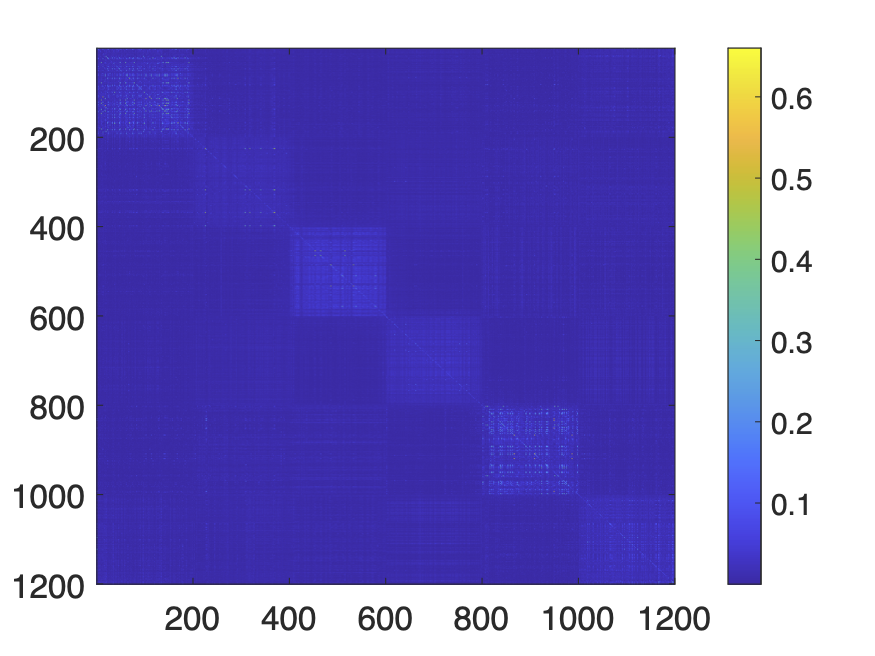}
\end{minipage}%
}%
\subfloat[ATTN-MSC]{
\begin{minipage}[t]{0.3\linewidth}
\centering
\includegraphics[scale=0.27]{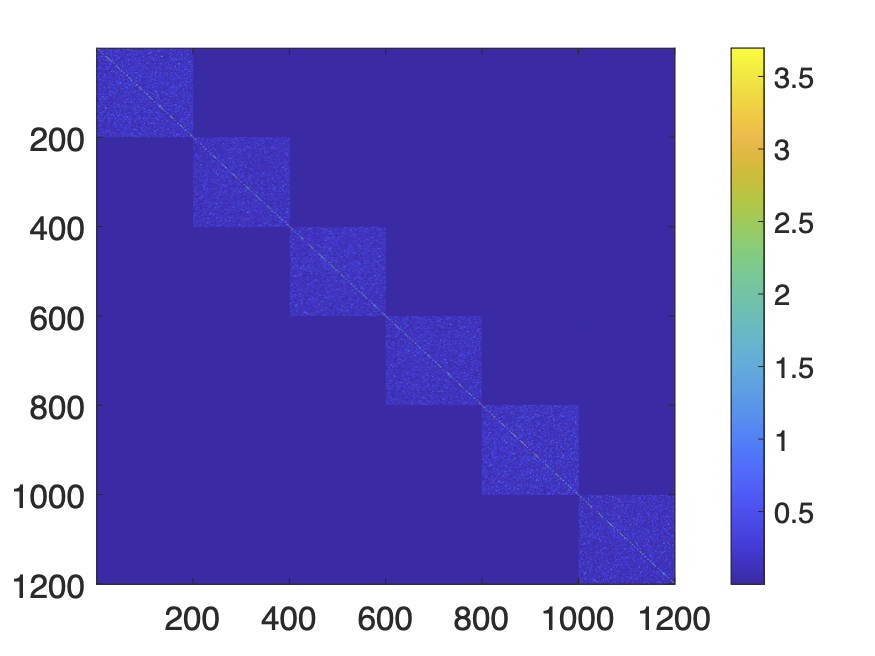}
\end{minipage}%
}%
\centering
\caption{The visualization of the affinity matrices about the Reuters datasets over the LTMSC, HLR-M$^{2}$SC and ATTN-MSC model.}
\label{fig:viamatrix}
\end{figure*}

\begin{figure*}[htbp]
\centering
\subfloat[Yale]{
\begin{minipage}[t]{0.3\linewidth}
\centering
\includegraphics[scale=0.27]{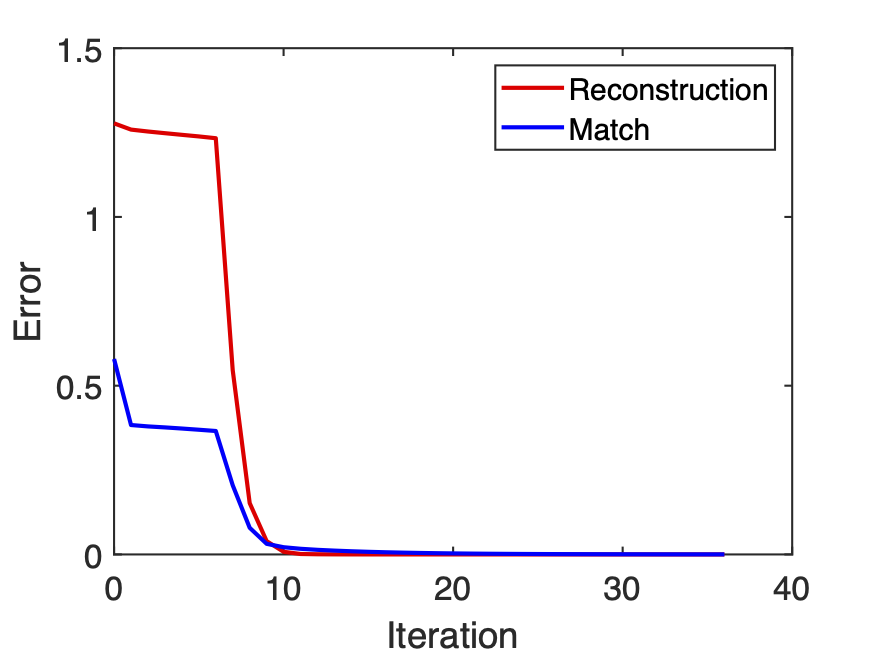}
\end{minipage}%
}%
\subfloat[MSRCV1]{
\begin{minipage}[t]{0.3\linewidth}
\centering
\includegraphics[scale=0.27]{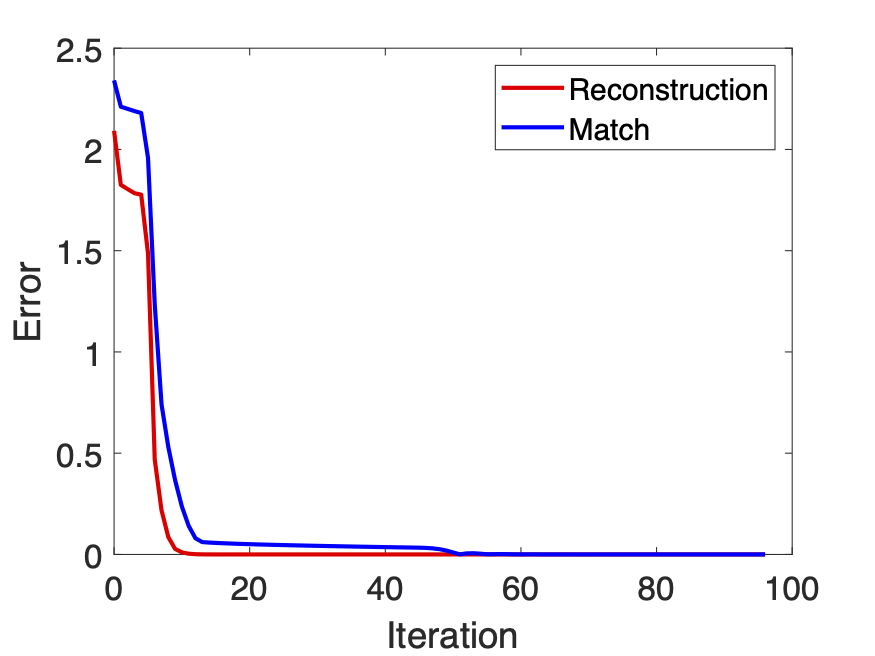}
\end{minipage}%
}%
\subfloat[100leaves]{
\begin{minipage}[t]{0.3\linewidth}
\centering
\includegraphics[scale=0.27]{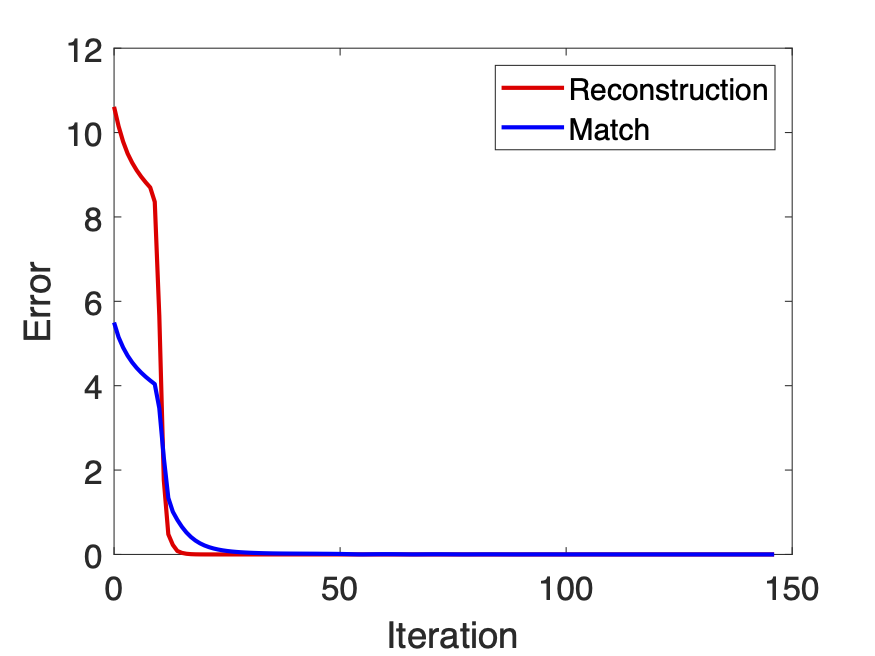}
\end{minipage}%
}%
\centering
\caption{Convergence results on the Yale, MSRCV1 and 100leaves datasets.}
\label{fig:convergence}
\end{figure*}

\begin{figure*}[htbp]
\centering
\subfloat[100leaves]{
\begin{minipage}[t]{0.33\linewidth}
\centering
\includegraphics[scale=0.17]{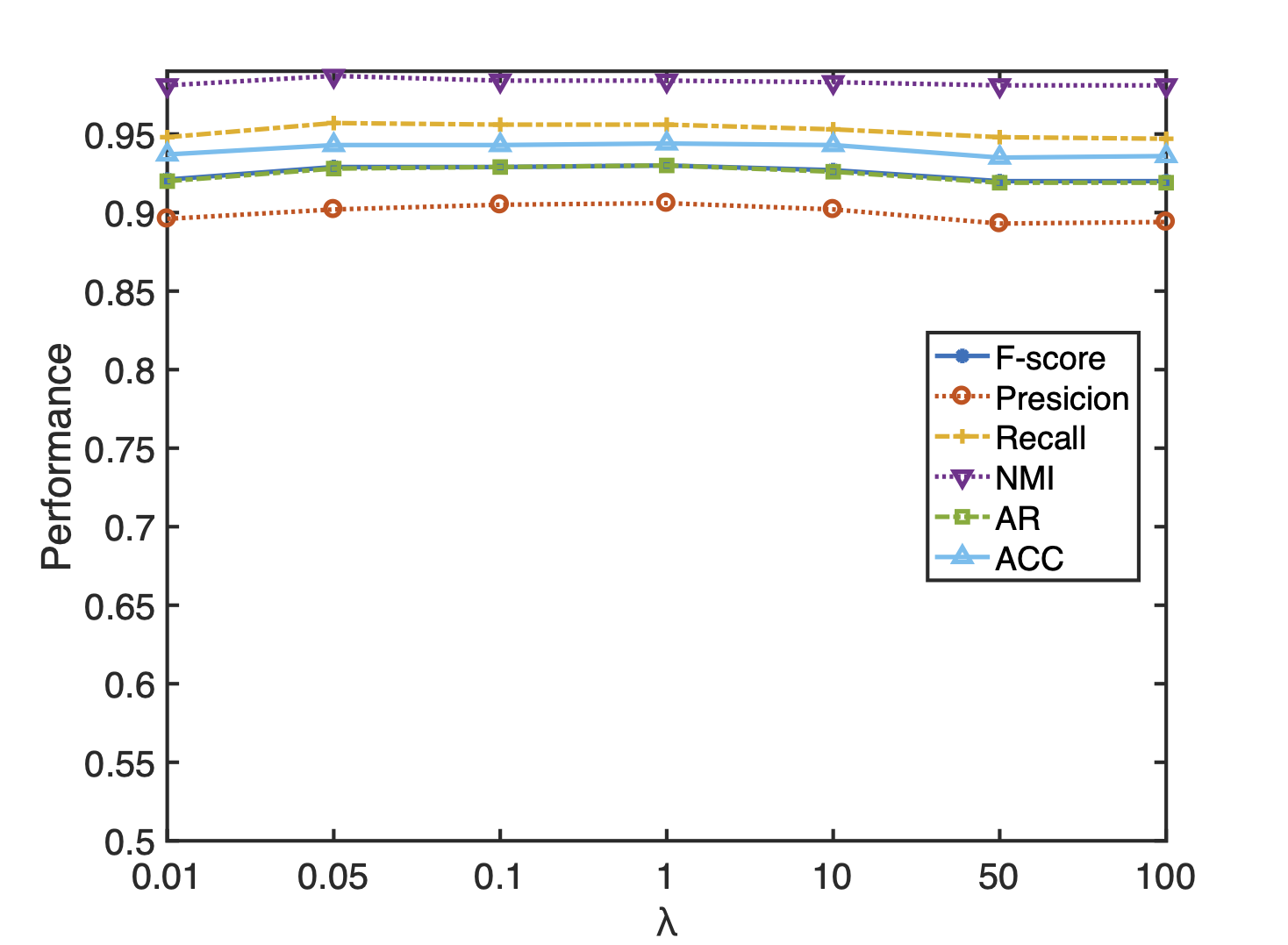}
\end{minipage}%
}%
\subfloat[MSRCV1]{
\begin{minipage}[t]{0.33\linewidth}
\centering
\includegraphics[scale=0.17]{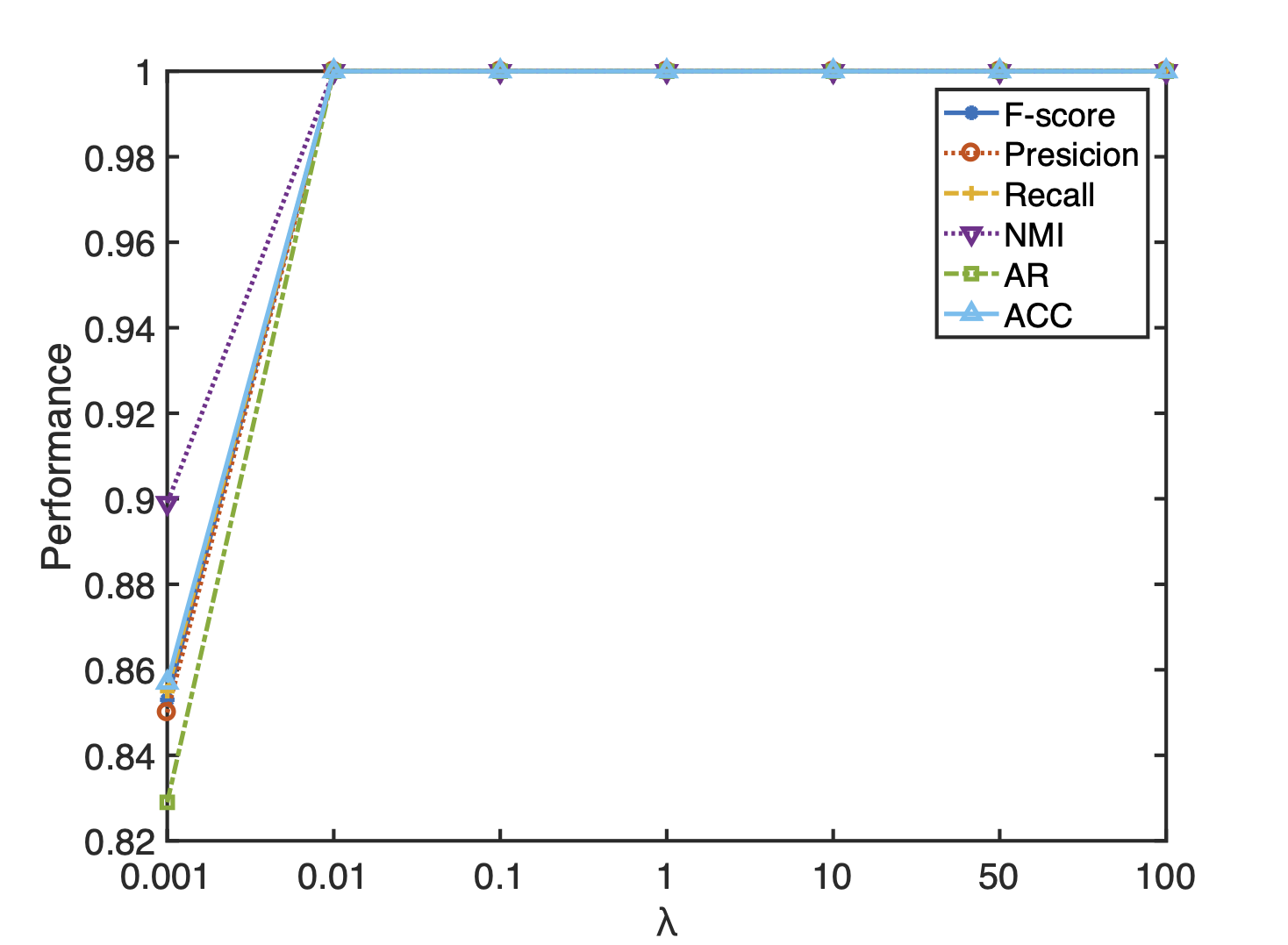}
\end{minipage}%
}%
\subfloat[Reuters]{
\begin{minipage}[t]{0.33\linewidth}
\centering
\includegraphics[scale=0.17]{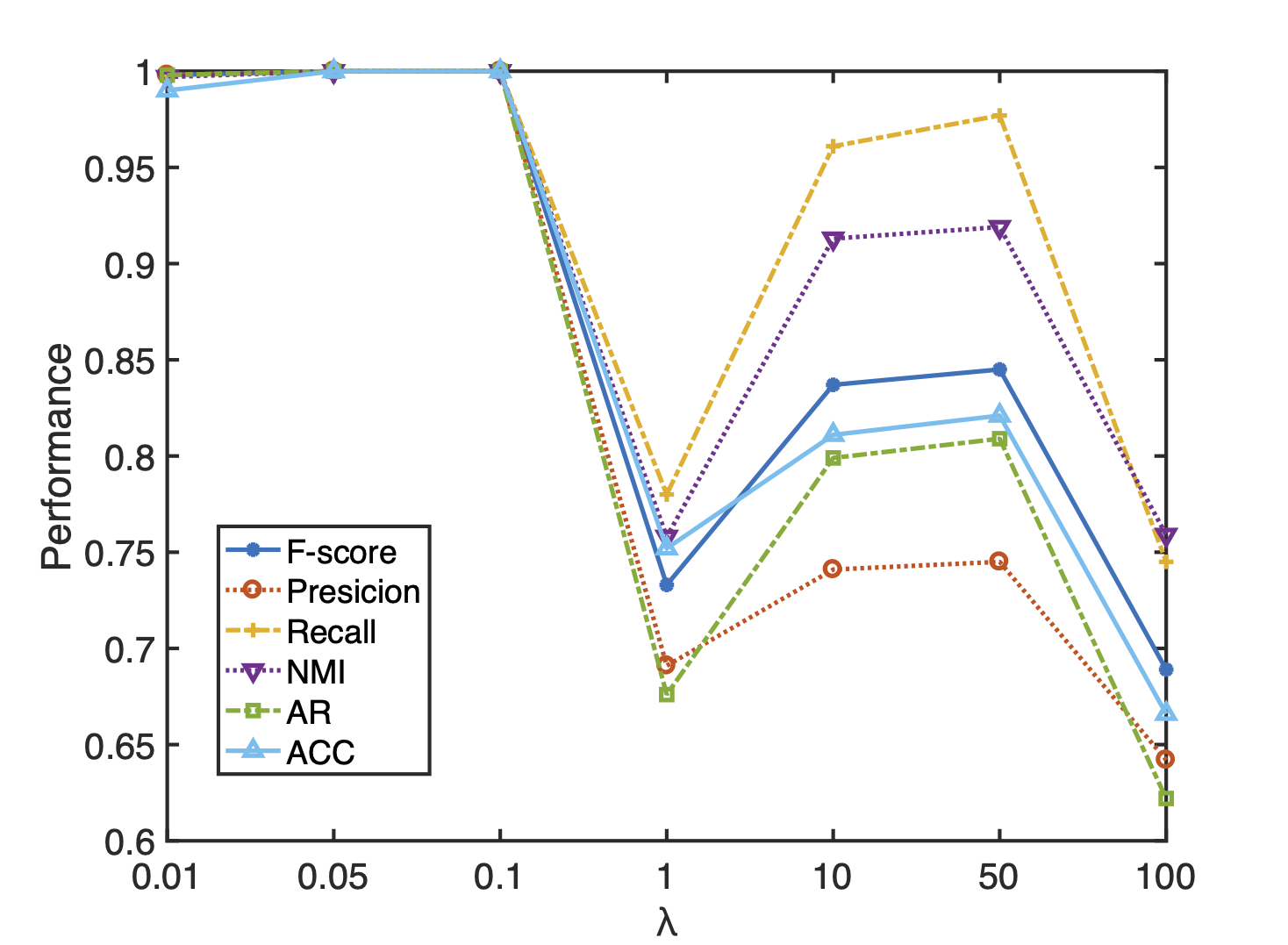}
\end{minipage}%
}%
\centering
\caption{Parameter tuning with respect to $\lambda$ on 100leaves, MSRCV1 and Reuters datasets.}
\label{fig:lam}
\end{figure*}

\begin{figure*}[htbp]
\centering
\includegraphics[scale=0.35]{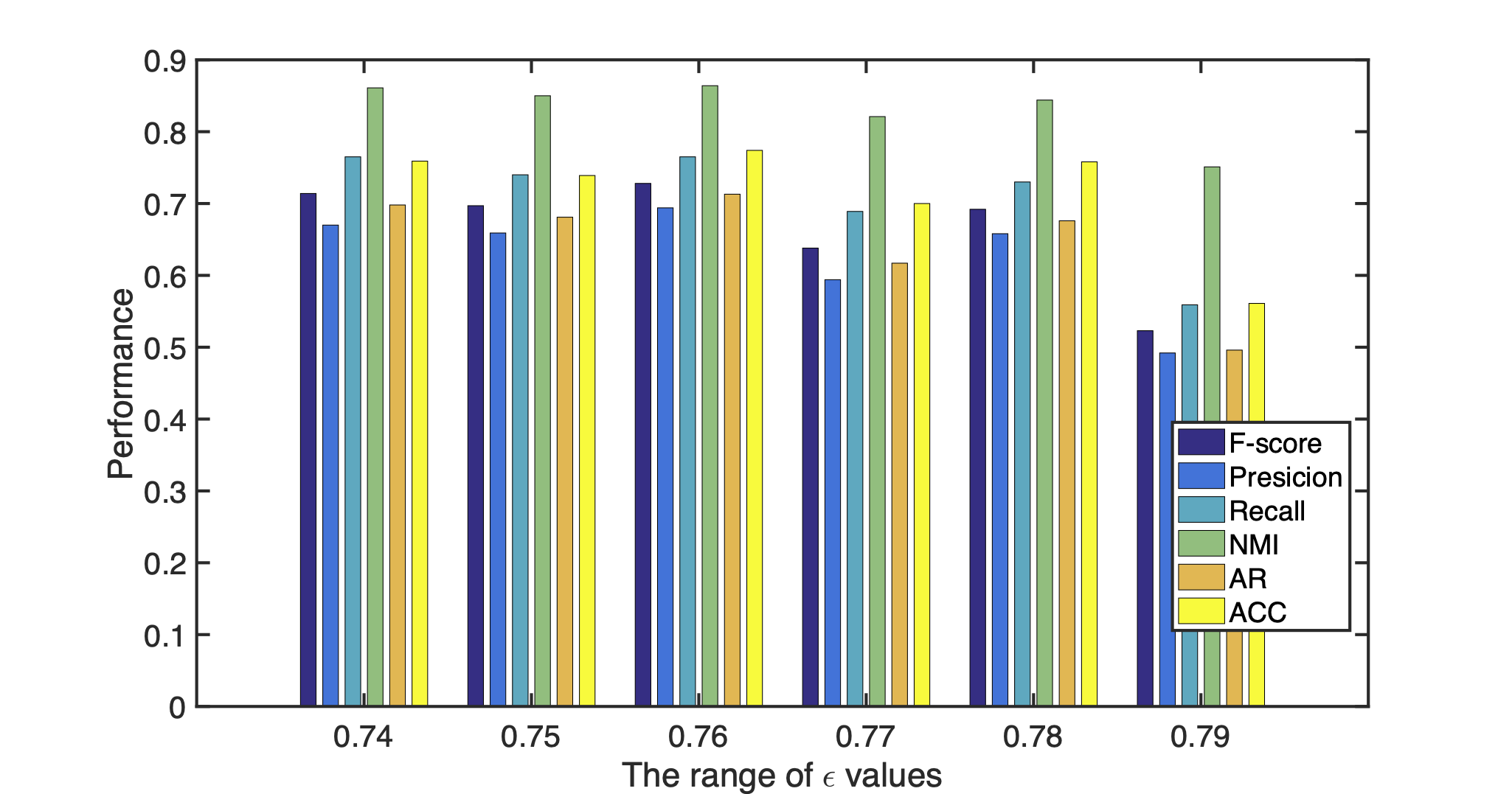}
\caption{Parameter tuning with respect to $\epsilon$ on Caltech101-7 dataset.}
\label{fig:calrse}
\end{figure*}

To verify its superiority, we compare the ATTN-MSC with two single-view and eleven multi-view methods: SPC$_{best}$~\cite{ng2002spectral}, LRR$_{best}$~\cite{liu2012robust}, Co-reg~\cite{kumar2011co}, DiMSC~\cite{cao2015diversity}, RMSC~\cite{xia2014robust}, LMSC~\cite{zhang2017latent}, ECMSC~\cite{wang2017exclusivity}, LTMSC~\cite{zhang2015low}, t-SVD-MSC~\cite{xie2018unifying}, ETLMSC~\cite{wu2019essential}, HLR-M$^{2}$VS~\cite{xie2020hyper}, CoMSC~\cite{liu2021multiview}, and {JSMC}~\cite{cai2023seeking}. The following state-of-the-art models are compared with TNAL-MSC regarding the clustering performance. Furthermore, detailed descriptions of those datasets and comparative approaches have been illustrated in supplementary materials.

\subsection{Evaluation Metrics}
In our experiments, six widely applied metrics will be utilized to evaluate the clustering performance: F-score, Precision, Recall, normalized mutual information (NMI), adjusted rand index (AR), and accuracy (ACC). The detailed information please refer to~\cite{lin2018multi,zhang2021joint,huang2019ultra,schutze2008introduction}. Obviously, the larger values of these metrics denote better clustering performance. 

\subsection{Clustering Performance Comparison}
Since the different initialization may result in different performances, in this paper, we report the average performance with the standard derivations of the compared algorithms, i.e., mean (standard deviation), by running 10 trials for each experiment. Apart from that, the best results will be highlighted in \textbf{boldface}, and the second-best results are $\underline{\text{underlined}}$. The detailed clustering results on six multi-view datasets have been carried out in Table \ref{tab:yale} and supplementary materials. From those experiment results, we can make the following observations.

In general, our proposed ATTN-MSC algorithm has achieved the best results under the six evaluation metrics on six multi-view datasets. Specifically, the proposed ATTN-MSC and t-SVD-MSC have reached the top two clustering performances on Yale, UCI Digits, and 100leaves datasets in almost all metrics. However, ATTN-MSC has improved the six metrics over t-SVD-MSC by $10.86\%$, $12.23\%$, $9.29\%$, $5.71\%$, $11.61\%$, and $7.07\%$ on the Yale dataset, respectively. On the UCI Digits datasets, ATTN-MSC has reached 1 in all six metrics, which is the same as ETLMSC, and has improved the six metrics over t-SVD-MSC by $5.26\%$, $5.26\%$, $5.15\%$, $5.60\%$, $5.82\%$, and $2.56\%$. As for the 100leaves dataset, ATTN-MSC has improved the clustering performance over t-SVD-MSC by $4.13\%$, $7.27\%$, $1.04\%$, $0.61\%$, $4.24\%$, and $4.98\%$ in F-score, Precision, Recall, NMI, AR and ACC, respectively. Apart from that, The most competitive model is the HLR-M$^{2}$VS in Caltech101-7 and MSRCV1 datasets. Nonetheless, our results are $8.73\%$, $0.96\%$ higher than HLR-M$^{2}$VS on average over these two datasets, respectively. Furthermore, on the Reuters dataset, ATTN-MSC improves around the $11.36\%$, $12.23\%$, $10.25\%$, $11.61\%$, $14.03\%$, and $8.70\%$ concerning six metrics over the ETLMSC method. In particular, CoMSC has achieved good performance in both 100leaves and Caltech101-7 datasets regarding the Recall metric. However, combining the other five clustering metrics, ATTN-MSC has improved $4.82\%$ and $24.61\%$ over CoMSC in total, showing advanced performance in the unbalanced dataset. Therefore, thanks to the promising capacity of capturing the low-rank information with the data-dependent framework of ATTNA, our proposed ATTN-MSC model is able to present better performance.
\begin{equation}
\begin{aligned}
\text {Reconstruction Error}&=\max\bigg\{\left\|\mathbf{X}_{v}-\mathbf{X}_{v} \mathbf{S}_{v}-\mathbf{E}_{v}\right\|_{\infty}\bigg\},\\
\text {Match Error}&=\max\bigg\{\left\|\mathbf{Z}_{v}-\mathbf{S}_{v}\right\|_{\infty}\bigg\},v=1,\cdots,V.
\end{aligned}
\label{eq:rematch}
\end{equation}

There are two main free parameters in our model: the balanced parameter $\lambda$, and the $\epsilon$. Due to the limitations of the paper, we only show the clustering performance results on 100leaves, MSRCV1 and Reuters datasets by the different $\lambda$ in Fig.~\ref{fig:lam}. It can be easily seen that the performance of ATTN-MSC is stable when $\lambda=\{0.01,0.05,0.1,1,10,50,100\}$, $\lambda=\{0.01,1,10,50,100\}$, and $\lambda=\{0.01,0.05,0.1\}$ on 100leaves, MSRCV1 and Reuters datasets, respectively. Besides, the evaluation results on Caltech101-7 dataset have been presented in Fig.~\ref{fig:calrse} with different $\epsilon$ tuning. Therefore, we can observe that ATTN-MSC can achieve promising performance while the pre-defined $\epsilon$ is set to 0.76.

Multi-view clustering models, such as Co-reg, DiMSC, RMSC, LMSC, ECMSC, CoMSC and JMSC, aim to harness the complementary information across multiple views. Therefore, Compared with the single-view based clustering methods $\text{SPC}_{best}$ and $\text{LRR}_{best}$, the multi-view methods have all achieved significant improvements. Furthermore, LTMSC, t-SVD-MSC, ETLMSC, HLR-M$^{2}$VS and ATTN-MSC are tensor based methods, whose clustering performance is almost higher than other matrix based methods. Moreover, it can be clearly seen that the ATTN-MSC is superior to the existing tensor-based methods on most real-world datasets, demonstrating that it is necessary to search the data-dependent optimal tensor network architecture.    

Furthermore, in Fig.~\ref{fig:conmatrix}, we present the confusion matrices of the top two methods on the Yale dataset. Compared to the t-SVD-MSC, the ATTN-MSC generates an improvement in almost all classes in terms of accuracy, which can be attributed to the effectiveness of optimal tensor network constraints. Besides, Fig.~\ref{fig:viamatrix} visualizes the affinity matrices captured by LTMSC, HLR-M$^{2}$VC, ATTN-MSC methods on the Reuters dataset.

\subsection{Model Discussion}
In Fig.~\ref{fig:convergence}, we plot the convergence curves of ATTN-MSC on Yale, MSRCV1 and 100leaves datasets, where the X-axis is the number of iterations, and Y-axis denotes the value of reconstruction error and match error (defined in Eq. (\ref{eq:rematch})) in each iteration step, respectively. Empirically, the number of optimization iteration with our method usually locates within the range of 40-150.

\section{Conclusions}
\label{sec:6}
We propose the optimally topological tensor network to describe different low rank information of self-representation tensors from various multi-view datasets. Specifically, based on the general tensor network, we automatically recognize the weakly correlated or uncorrelated correlations and prune them. The corresponding new structure is fed into greedy-based rank incremental algorithm to achieve the optimal tensor network finally. To demonstrate our proposed model's significant performance, we employ it in one popular application, i.e., multi-view subspace clustering. Extensive experimental results on six real-world datasets have illustrated the superiority of our model ATTN-MSC compared to the multiple state-of-art clustering methods. In the future, it is promising to apply it to other applications such as high-order tensor completion, image recovery, etc.

  \section*{Acknowledgments}

This work was supported by the National Natural Science Foundation of China (NSFC) under Grant 62171088, Grant 6220106011 and Grant U19A2052.

\bibliographystyle{plain}
\bibliography{cite}

%
\clearpage

\appendix

 \section*{Supplementary Material: Adaptively Topological Tensor Network for Multi-view Subspace Clustering}



\section{An Example of ATTN}

\begin{figure}[htbp]
\centering
\includegraphics[scale=0.45]{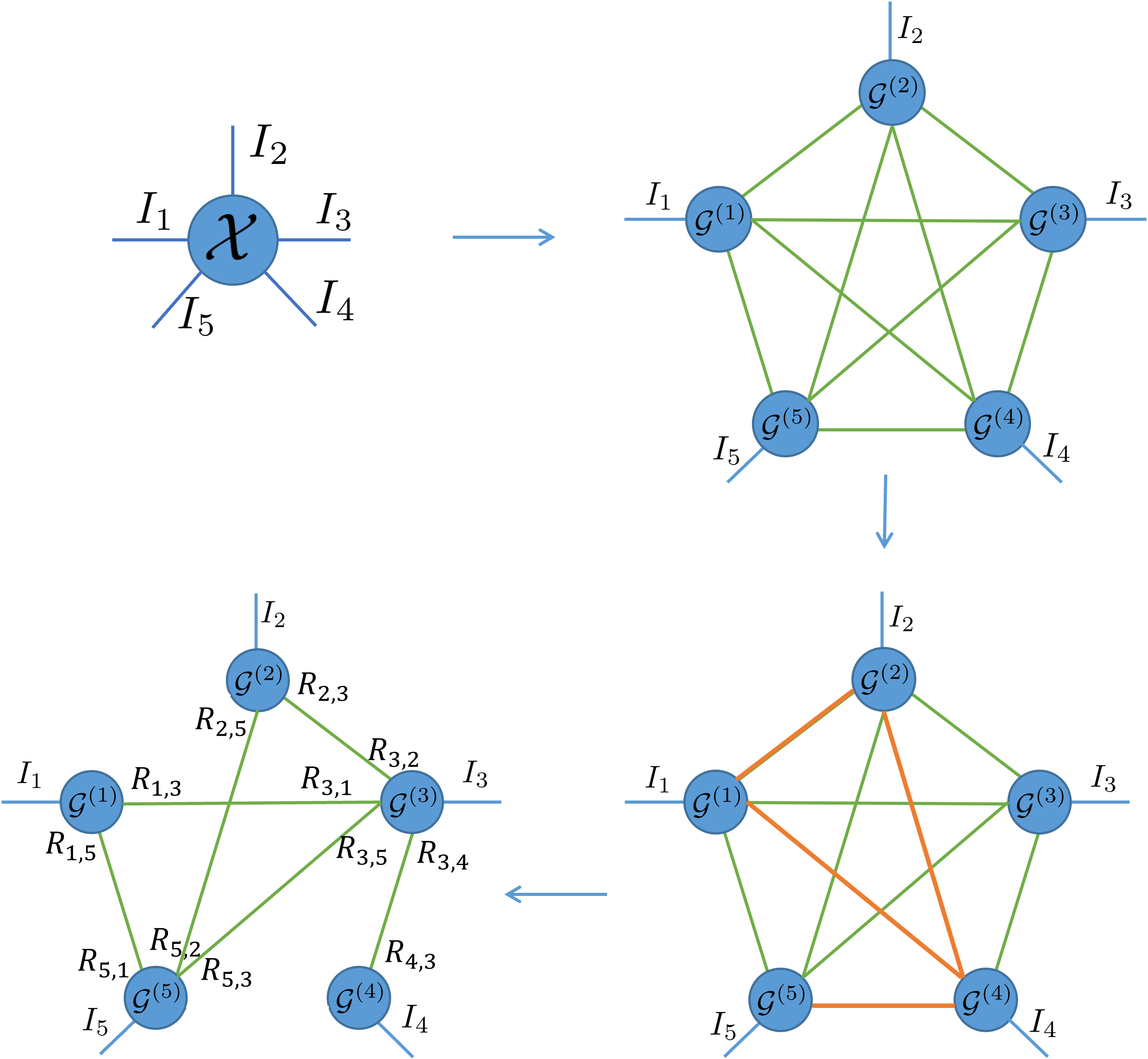}
\caption{The graphic illustration of generating a tensor newtork topology of the 5th-order tensor $\mathcal{X}^{I_{1} \times I_{2} \times I_{3} \times I_{4} \times I_{5}}$.}
\label{fig:dtn}
\end{figure}

An example of the process of searching the specific tensor network topology for a 5th-order tensor has shown in Fig. \ref{fig:dtn}. Specifically, the self-representation tensor is firstly represented by an existing complex tensor network, e.g., FCTN. Based on that, the connections in the FCTN are judged to be redundant according to their correlations. If there is no or weak correlation between tensor network adjacent factors, the ties of these factors (the red lines in Fig. \ref{fig:dtn}) will be clipped. 

\section{The overview of The ATTN Algorithm}
The optimization procedures of ATTN are outlined in Algorithm \ref{alg:prune1} and Algorithm \ref{alg:prune3}.

\begin{algorithm}[htbp]
	\caption{ATTN Algorithm (ATTNA)}
		\begin{algorithmic}[1]
			\STATE \textbf{Input:} An $N$th-order data tensor $\mathcal{X} \in \mathbb{R}^{I_1 \times I_2 \times \cdots \times I_N}$, the initial ranks $r = 2$, maximum number of iterations $\text{iter}_{\text{max}}$, the threshold for stopping the algorithm $\text{tol}_{\text{als}}=10^{-6}$,  predefined accuracy $\varepsilon$, $\text{iter}=1$, and $B$. 
			\STATE \textbf{Initialize:}  Random initialization $\mathcal{G}^{(1)}, \cdots,\mathcal{G}^{(N)}$.\\
			\STATE $\mathcal{X}_{\text{new}}=\operatorname{TC}(\mathcal{G}^{(1)}, \cdots,\mathcal{G}^{(N)})$;
			 \STATE \textbf{while} $\text{iter} \leq \text{iter}_{\max }$ \textbf{do}
			 \STATE $\mathcal{X}_{\text{last}}=\mathcal{X}_{\text{new}}$;
			 \STATE  ~~\textbf{for} $n=1,\cdots,N$ \textbf{do}
			 \STATE~~~~~~$\mathcal{A}^{\neq n} =\operatorname{TC}(\mathcal{G}^{(1)}, \cdots,\mathcal{G}^{(n-1)}, \mathcal{G}^{(n+1)},\cdots,\mathcal{G}^{(N)})$;
			 \STATE~~~~~~$\mathbf{A}^{\neq n}_{(n)}$=$\operatorname{reshape}(\mathcal{A}^{\neq n},[~], I_1 \cdots I_N/I_n$);
			 \STATE~~~~~~$\mathbf{G}_{(1)}^{(n)}\leftarrow \arg \min \left\|\mathbf{X}_{(n)}-(\mathbf{G}_{(1)}^{(n)})(\mathbf{A}^{\neq n}_{(n)})\right\|_{\mathrm{F}}$;
			 \STATE~~~~~~$\mathcal{G}^{(n)}$=$\operatorname{reshape}(\mathbf{G}_{(1)}^{(n)}$, $\operatorname{size}(\mathcal{G}^{(n)}$));
		     \STATE~~\textbf{end for}
			 \STATE $\mathcal{X}_{\text{new}}=\operatorname{TC}(\mathcal{G}^{(1)}, \cdots,\mathcal{G}^{(N)})$;
			 \STATE  \textbf{if} $\frac{\left\|\mathcal{X}_{\text{last}}-\mathcal{X}_{\text{new}}\right\|_{\mathrm{F}}}{\left\|\mathcal{X}_{\text{last}}\right\|_{\mathrm{F}}} \leq \text{tol}_{\text {als }}$ \textbf{then}
			 \STATE ~~~~break;
			 \STATE \textbf{end if}
			 \STATE $\text{iter}=\text{iter}+1$;
			 \STATE \textbf{end while}
		     \STATE \textbf{for} $b=1,\cdots,B$ \textbf{do}
		     \STATE ~~~~~Calculate $\operatorname{\delta}^{(b)} = \operatorname{RSE}^{\Psi-R_{i,j}} - \operatorname{RSE}^{\Psi}$;
		     \STATE \textbf{end for}
		     \STATE Select the redundant connections, and set the corresponding edge ranks to 1;
			\STATE Optimize the new tensor network by additionally applying the rank incremental strategy with a greedy algorithm, i.e., Algorithm \ref{alg:prune3}.
			\STATE \textbf{Output:} $\mathcal{G}^{(1)}, \cdots,\mathcal{G}^{(N)}$, $\mathcal{X}_{\text{new}}$
	\end{algorithmic}
	\label{alg:prune1}
\end{algorithm}

 \begin{algorithm}[htbp]
 	\caption{Rank incremental algorithm with greedy algorithm}
 		\begin{algorithmic}[1]
 			\STATE\textbf{Input:} $\mathcal{G}^{(1)}, \cdots,\mathcal{G}^{(N)}$, $\mathcal{X} \in \mathbb{R}^{I_1 \times I_2 \times \cdots \times I_N}$, edge ranks $\mathbf{R}_{\text{OA}}$, threshold $\varepsilon$, $\text{iter}_{\text{max}}$, $i=1$, $\text{step size} = a$;
 			\STATE \textbf{while} $i \leq \text{iter}_\text{max}$ \textbf{do}
 			\STATE \quad Repeat 6-12 lines in ATTNA;
 			\STATE \quad \textbf{if} $\frac{\left\|\mathcal{X}_{\text{new}}-\mathcal{X}\right\|_{\mathrm{F}}}{\left\|\mathcal{X}\right\|_{\mathrm{F}}} > \varepsilon$ and \textbf{then}
 			\STATE \quad \quad Select the edge rank $R_{i,j}$;
 			\STATE \quad \quad $R_{i,j} = R_{i,j}+1$;
 			\STATE \quad \quad Add a new slice to $\mathcal{G}^{(i)}$;
 			\STATE \quad \quad Add a new slice to $\mathcal{G}^{(j)}$;
 			\STATE \quad \textbf{end if}
 			\STATE \quad i = i + a;			
 			\STATE \textbf{end while}
 			\STATE\textbf{Output:} $\mathcal{G}^{(1)}, \cdots,\mathcal{G}^{(N)}$.
 	\end{algorithmic}
 	\label{alg:prune3}
 \end{algorithm}

\section{Experimental Relevant Materials}

\subsection{The image samples in image reconstruction}
\begin{figure}[htbp]
\centering
\centering
\includegraphics[scale=0.45]{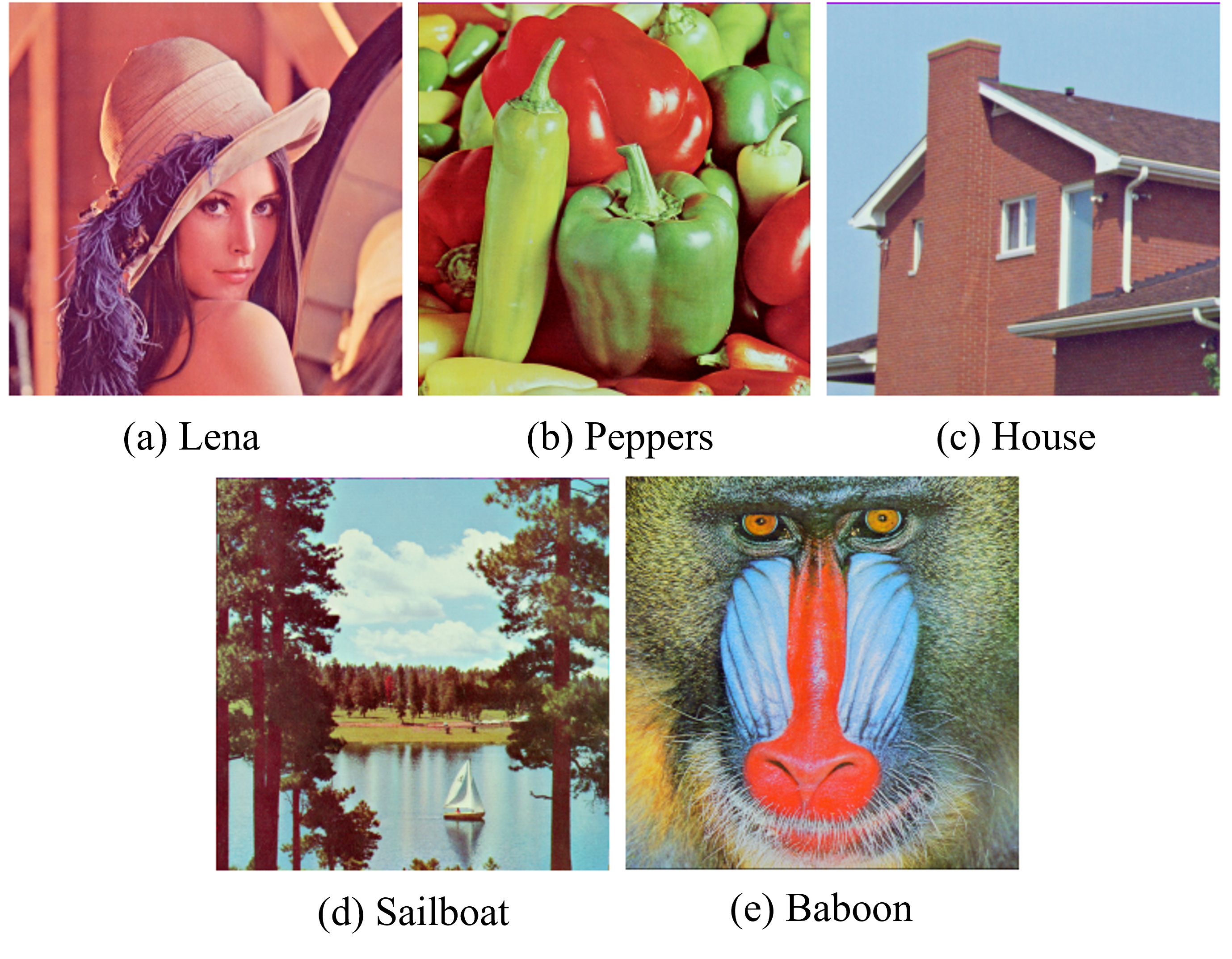}
\caption{The five images in image reconstruction experiments.}
\label{fig:picreconstruct}
\end{figure}

Five real-world images have been utilized to demonstrate the superiority of our proposed tensor network framework by conducting image reconstruction experiments. As shown in Fig. \ref{fig:picreconstruct}, they are renamed "Lena", "Peppers", "House", "Sailboat" and "Baboon" in this paper.

\subsection{The Multi-view Datasets Description}
\textbf{Yale}\footnote{http://cvc.cs.yale.edu/cvc/projects/yalefaces/yalefaces.html} face dataset contains 165 grayscale images of 15 individuals. Each class is given 11 images, one of which corresponded to a different facial expression or conﬁguration. We extracted raw pixels of dimension 4096, the LBP feature of dimension 3304 and the Gabor feature of dimension 6750 as 3 views for our experiments.

\textbf{UCI Digit}\footnote{http://archive.ics.uci.edu/ml} face dataset contains 2000 grayscale images of handwritten digits of 0 to 9. Each digit contains 200 samples. Three features are selected, including pixel averages in 2 × 3 windows (PIX) of dimension 240, Fourier coefficients of dimension 76 and morphological features (MOR) of dimension 6.

\textbf{Reuters}\footnote{http://lig-membres.imag.fr/grimal/data.html} contains 2000 documents from six types, each of which is described in five languages, including English, French, German, Italian, and Spanish. The dimensions of these views are 4819, 4810, 4892, 4858 and 4777, respectively.

\textbf{One-hundred plant species leaves dataset (100leaves)}\footnote{https://archive.ics.uci.edu/ml/datasets/One-hundred+plant+species+leaves+data+set} generated from the UCI repository. This multi-view dataset consists of 1600 samples with three views, including shape descriptors, fine-scale boundaries and texture information. The dimension of these views is all 64. Each sample is one of the one hundred plant species \cite{wang2019study}.

\textbf{Caltech101-7}\footnote{http://www.vision.caltech.edu/Image\_\text{Datasets/Caltech101/}} is a subset of Caltech101, whose per category contains about 40 to 800 images. In this dataset, seven popular courses, including faces, motorcycles, dollar bills, Garfield, Snoopy, stop signs and Windsor chairs, have been selected \cite{liu2021multiview}. We select six views, including a 48-dimension Gabor feature, a 40-dimension wavelet moments (Wm), a 254-dimension centrist feature, a 1984-dimension HOG feature, a 512-dimension Gist feature, and a 928-dimension LBP feature. 

\textbf{MSRCV1}\footnote{http://research.microsoft.com/en-us/projects/objectclassrecognition/} is a scene dataset, which consists of 210 image samples collected from 7 clusters with 6 views, including CENTRIST of dimension 1302, CMT of dimension 48, GIST of dimension 512, HOG of dimension 100, LBP of dimension 256, and DoG-SIFT of dimension 210.

\subsection{The Description Of The Ten Comparative Approach}
\textbf{SPC$_{best}$}~\cite{ng2002spectral}: obtain the best clustering segmentation among all views with the standard spectral clustering.

\textbf{LRR$_{best}$}~\cite{liu2012robust}: obtain the best clustering segmentation among all views with the low-rank representation.

\textbf{Co-reg}~\cite{kumar2011co}: proposes the co-regularization schemes, which are employed in a spectral clustering algorithm to attain the clustering results.

\textbf{DiMSC}~\cite{cao2015diversity}: utilizes the Hilbert-Schmidt Independence Criterion to be a diversity term that enforces the learned self-representations different from each other.

\textbf{RMSC}~\cite{xia2014robust}: is a novel Markov chain method for robust multi-view spectral clustering via recovering a shared low-rank transition probability matrix.

\textbf{LMSC}~\cite{zhang2017latent}: is a latent multi-view subspace clustering model which seeks the underlying latent representation and exploits the complementary information across the multiple views simultaneously.

\textbf{ECMSC}~\cite{wang2017exclusivity}: is a novel multi-view subspace clustering method, which introduces a position-aware exclusivity term and a consistency term to enhance the performance of clustering.

\textbf{LTMSC}~\cite{zhang2015low}: adopts the low-rank tensor constraint over the representation matrices for multi-view subspace clustering.

\textbf{t-SVD-MSC}~\cite{xie2018unifying}: uses the t-SVD based tensor nuclear norm to learn the optimal subspace based on self-representation assumptions.

\textbf{ETLMSC}~\cite{wu2019essential}: is an essential tensor learning method based on the Markov chain to explore the high-order correlations for multi-view representations.

\textbf{HLR-M$^{2}$VS}~\cite{xie2020hyper}: deals with the multi-view nonlinear subspace representation problem based on t-SVD.

\textbf{CoMSC}~\cite{liu2021multiview}: develops an elegant MSC model to group the data and remove the redundancy information concurrently.

\textbf{JSMC}~\cite{cai2023seeking}: focuses on seeking both commonness and inconsistencies across different views in the subspace representation learning.

\subsection{Experiment Result About Multi-view Subspace Clustering}
The clustering results on UCI Digits, Reuters, 100leaves, Caltech101-7, and MSRCV1 have been shown from Table \ref{tab:UCI} to Table \ref{tab:msrcv1}. The best results will be highlighted in \textbf{boldface} and the second-best results are $\underline{\text{underlined}}$.

\begin{table*}[htbp]
\begin{center}
\caption{Clustering results on UCI Digits dataset. For ATTN-MSC, we set $\lambda=0.1$.}
\label{tab:UCI}
\scalebox{0.6}{
\begin{tabular}{{cccccccccc}}
\toprule		
&Method &F-score &Precision &Recall &NMI &AR &ACC\\
\midrule
	&$\text{SPC}_{best}$ 	&0.517(0.001)	&0.508(0.001)	&0.902(0.000)	&0.576(0.000)	&0.462(0.001)	&0.644(0.002)\\		
	&$\text{LRR}_{best}$	&0.441(0.003)	&0.434(0.003)	&0.449(0.004)	&0.507(0.003)	&0.378(0.004)	&0.571(0.002)\\
\midrule
	&Co-reg		&0.705(0.033)	&0.694(0.040)	&0.715(0.027)	&0.732(0.023)	&0.671(0.037)	&0.804(0.048)\\
	&DiMSC		&0.308(0.009)	&0.276(0.009)	&0.346(0.009)	&0.391(0.011)	&0.221(0.010)	&0.453(0.026)\\
	&RMSC 		&0.845(0.050)	&0.842(0.055)	&0.848(0.046)	&0.850(0.031)	&0.828(0.056)	&0.903(0.055)\\
	&LMSC 		&0.744(0.001)	&0.742(0.001)	&0.746(0.001)	&0.776(0.001)	&0.716(0.001)	&0.846(0.001)\\
	&ECMSC		&0.709(0.004)	&0.661(0.003)	&0.764(0.006)	&0.781(0.003)	&0.674(0.005)	&0.719(0.006)\\
 	&LTMSC 		&0.740(0.015)	&0.727(0.014)	&0.753(0.016)	&0.765(0.012)	&0.710(0.017)	&0.794(0.011)\\
	&t-SVD-MSC 	&0.950(0.000)	&0.950(0.000)	&0.951(0.000)	&0.947(0.000)	&0.945(0.000)	&0.975(0.000)\\
	&ETLMSC	&\textbf{1.000(0.000)}	&\textbf{1.000(0.000)}	&\textbf{1.000(0.000)}	&\textbf{1.000(0.000)}	&\textbf{1.000(0.000)}	&\textbf{1.000(0.000)}\\
	&HLR-M$^{2}$VS	&\underline{0.996(0.000)}	&\underline{0.996(0.000)}	&\underline{0.996(0.000)}	&\underline{0.995(0.000)}	&\underline{0.996(0.000)}	&\underline{0.998(0.000)}\\
    &CoMSC &0.660(0.000)	&0.648(0.000)	&0.931(0.000)	&0.707(0.000)	&0.622(0.000)	&0.748(0.001)\\
    &JSMC &0.762(0.000)	&0.726(0.000)	&0.802(0.000)	&0.804(0.000)	&0.734(0.000)	&0.796(0.000)\\
	&ATTN-MSC	&\textbf{1.000(0.000)}	&\textbf{1.000(0.000)}	&\textbf{1.000(0.000)}	&\textbf{1.000(0.000)}	&\textbf{1.000(0.000)}	&\textbf{1.000(0.000)}\\
\bottomrule
\end{tabular}
}
\end{center}
\end{table*}

\begin{table*}[htbp]
\begin{center}
\caption{Clustering results on Reuters dataset. For ATTN-MSC, we set $\lambda=0.1$.}
\label{tab:reuters}
\scalebox{0.6}{
\begin{tabular}{{cccccccccc}}
\toprule		
&Method &F-score &Precision &Recall &NMI &AR &ACC\\
\midrule
	&$\text{SPC}_{best}$ 	&0.344(0.003)	&0.317(0.017)	&0.379(0.032)	&0.267(0.008)	&0.199(0.010)	&0.473(0.021)\\		
	&$\text{LRR}_{best}$	&0.297(0.000)	&0.241(0.000)	&0.387(0.000)	&0.184(0.000)	&0.116(0.000)	&0.376(0.000)\\
\midrule
	&Co-reg		&0.347(0.004)	&0.319(0.013)	&0.383(0.020)	&0.275(0.007)	&0.203(0.009)	&0.466(0.003)\\
	&DiMSC		&0.352(0.000)	&0.309(0.000)	&0.408(0.000)	&0.285(0.000)	&0.201(0.000)	&0.462(0.000)\\
	&RMSC 		&0.398(0.007)	&0.382(0.012)	&0.416(0.012)	&0.348(0.013)	&0.272(0.010)	&0.542(0.024)\\
	&LMSC 		&0.407(0.000)	&0.374(0.000)	&0.447(0.000)	&0.342(0.000)	&0.276(0.000)	&0.562(0.000)\\
	&ECMSC		&0.279(0.000)	&0.200(0.000)	&0.461(0.000)	&0.130(0.000)	&0.061(0.000)	&0.315(0.000)\\
 	&LTMSC 		&0.304(0.000)	&0.253(0.000)	&0.380(0.000)	&0.196(0.001)	&0.130(0.000)	&0.382(0.001)\\
	&t-SVD-MSC 	&0.792(0.001)	&0.784(0.000)	&0.800(0.001)	&0.779(0.001)	&0.750(0.001)	&0.882(0.000)\\
	&ETLMSC	&\underline{0.898(0.114)}	&\underline{0.891(0.128)}	&\underline{0.907(0.098)}	&\underline{0.896(0.087)}	&\underline{0.877(0.138)}	&\underline{0.920(0.118)}\\
	&HLR-M$^{2}$VS	&0.714(0.001)	&0.704(0.001)	&0.724(0.001)	&0.708(0.001)	&0.656(0.001)	&0.831(0.001)\\
    &CoMSC &0.407(0.000)	&0.349(0.000)	&0.764(0.000)	&0.330(0.000)	&0.264(0.000)	&0.547(0.000)\\
    &JSMC &0.369(0.000)	&0.290(0.000)	&0.507(0.000)	&0.290(0.000)	&0.200(0.000)	&0.463(0.000)\\
	&ATTN-MSC	&\textbf{1.000(0.000)}	&\textbf{1.000(0.000)}	&\textbf{1.000(0.000)}	&\textbf{1.000(0.000)}	&\textbf{1.000(0.000)}	&\textbf{1.000(0.000)}\\
\bottomrule
\end{tabular}
}
\end{center}
\end{table*}

\begin{table*}[htbp]
\begin{center}
\caption{Clustering results on 100leaves dataset. For ATTN-MSC, we set $\lambda=0.1$.}
\label{tab:100leaves}
\scalebox{0.6}{
\begin{tabular}{{cccccccccc}}
\toprule		
&Method &F-score &Precision &Recall &NMI &AR &ACC
\\
\midrule
	&$\text{SPC}_{best}$ 		&0.215(0.008)	&0.128(0.007)	&0.674(0.006)	&0.777(0.002)	&0.203(0.008)	&0.483(0.014)\\			
	&$\text{LRR}_{best}$		&0.315(0.010)	&0.274(0.007)	&0.371(0.008)	&0.715(0.018)	&0.307(0.011)	&0.488(0.013)\\
\midrule
	&Co-reg		&0.757(0.025)	&0.702(0.034)	&0.823(0.017)	&0.926(0.008)	&0.755(0.026)	&0.801(0.069)\\
	&DiMSC		&0.580(0.005)	&0.647(0.001)	&0.882(0.002)	&0.713(0.003)	&0.611(0.002)	&0.857(0.004)\\
	&RMSC 		&0.733(0.022)	&0.687(0.024)	&0.785(0.020)	&0.919(0.007)	&0.730(0.022)	&0.782(0.022)\\
	&LMSC 		&0.649(0.013)	&0.608(0.011)	&0.695(0.012)	&0.869(0.009)	&0.645(0.012)	&0.738(0.017)\\
	&ECMSC		&0.521(0.011)	&0.459(0.007)	&0.602(0.006)	&0.823(0.013)	&0.516(0.013)	&0.670(0.012)\\
 	&LTMSC 		&0.637(0.015)	&0.605(0.013)	&0.672(0.018)	&0.868(0.007)	&0.633(0.015)	&0.732(0.013)\\
	&t-SVD-MSC 	&\underline{0.921(0.013)}	&\underline{0.880(0.019)}	&\underline{0.966(0.008)}	&\underline{0.984(0.003)}	&\underline{0.920(0.013)}	&\underline{0.924(0.013)}\\
	&ETLMSC		&0.843(0.026)	&0.773(0.039)	&0.927(0.009)	&0.966(0.005)	&0.841(0.027)	&0.847(0.027)\\
	&HLR-M$^{2}$VS	&0.912(0.012)	&0.870(0.015)	&0.958(0.009)	&0.982(0.003)	&0.911(0.012)	&0.911(0.010)\\
	&CoMSC &0.895(0.000)	&0.865(0.000)	&\textbf{0.998(0.000)}	&0.968(0.000)	&0.894(0.000)	&0.921(0.000)\\
	&JSMC &0.667(0.000)	&0.607(0.000)	&0.741(0.000)	&0.883(0.000)	&0.664(0.000)	&0.731(0.000)\\
	&ATTN-MSC	&\textbf{0.929(0.007)}	&\textbf{0.905(0.011)}	&0.956(0.004)	&\textbf{0.984(0.001)}	&\textbf{0.929(0.007)}	&\textbf{0.943(0.008)}\\
\bottomrule
\end{tabular}
}
\end{center}
\end{table*}

\begin{table*}[htbp]
\begin{center}
\caption{Clustering results on Caltech101-7 dataset. For ATTN-MSC, we set $\lambda=0.01$.}
\label{tab:cal}
\scalebox{0.6}{
\begin{tabular}{{cccccccccc}}
\toprule		
&Method &F-score &Precision &Recall &NMI &AR &ACC
\\
\midrule
	&$\text{SPC}_{best}$ 	&0.447(0.021)	&0.764(0.025)	&0.316(0.017)	&0.429(0.035)	&0.286(0.024)	&0.447(0.035)\\			
	&$\text{LRR}_{best}$	&0.551(0.001)	&0.796(0.001)	&0.422(0.000)	&0.540(0.000)	&0.388(0.001)	&0.568(0.001)\\
\midrule
	&Co-reg		&0.329(0.017)	&0.511(0.015)	&0.243(0.016)	&0.180(0.010)	&0.108(0.015)	&0.339(0.007)\\
	&DiMSC		&0.365(0.003)	&0.639(0.006)	&0.255(0.002)	&0.262(0.004)	&0.185(0.004)	&0.384(0.003)\\
	&RMSC 		&0.452(0.018)	&0.797(0.028)	&0.316(0.014)	&0.469(0.004)	&0.299(0.021)	&0.427(0.016)\\
	&LMSC 		&0.350(0.001)	&0.633(0.002)	&0.242(0.001)	&0.248(0.001)	&0.174(0.001)	&0.350(0.001)\\
	&ECMSC		&0.533(0.011)	&0.756(0.008)	&0.411(0.010)	&0.539(0.010)	&0.358(0.012)	&0.536(0.011)\\
 	&LTMSC 		&0.562(0.004)	&0.877(0.003)	&0.414(0.003)	&0.591(0.007)	&0.418(0.004)	&0.567(0.000)\\
	&t-SVD-MSC 	&0.695(0.001)	&\underline{0.977(0.000)}	&0.540(0.001)	&0.762(0.001)	&0.580(0.001)	&0.663(0.000)\\
	&ETLMSC		&0.504(0.043)	&0.863(0.024)	&0.357(0.040)	&0.605(0.006)	&0.361(0.045)	&0.424(0.061)\\
	&HLR-M$^{2}$VS	&\underline{0.707(0.000)}	&\textbf{0.972(0.000)}	&0.555(0.000)	&\underline{0.764(0.000)}	&\underline{0.593(0.000)}	&\underline{0.690(0.000)}\\ 
	&CoMSC &0.660(0.000)	&0.717(0.000)	&\underline{0.757(0.000)}	&0.524(0.000)	&0.473(0.000)	&0.675(0.000)\\
 	&JSMC &0.624(0.000)	&0.884(0.000)	&0.483(0.000)	&0.569(0.000)	&0.484(0.000)	&0.615(0.000)\\
	&ATTN-MSC	&\textbf{0.728(0.015)}	&0.694(0.015)	&\textbf{0.765(0.015)}	&\textbf{0.864(0.007)}	&\textbf{0.713(0.016)}	&\textbf{0.774(0.022)}\\
\bottomrule
\end{tabular}
}
\end{center}
\end{table*}

\begin{table*}[t!]
\begin{center}
\caption{Clustering results on MSRCV1 dataset. For ATTN-MSC, we set $\lambda=0.1$.}
\label{tab:msrcv1}
\scalebox{0.6}{
\begin{tabular}{{cccccccc}}
\toprule		
&Method &F-score &Precision &Recall &NMI &AR &ACC
\\
\midrule
		&$\text{SPC}_{best}$		&0.598(0.062)	&0.592(0.066)	&0.605(0.060)	&0.626(0.053)	&0.533(0.073)	&0.710(0.078)\\
		&$\text{LRR}_{best}$ 		&0.537(0.006)	&0.530(0.007)	&0.544(0.005)	&0.571(0.004)	&0.462(0.007)	&0.676(0.006)\\
		\midrule
		&Co-reg 		&0.674(0.035)	&0.659(0.046)	&0.690(0.030)	&0.694(0.027)	&0.619(0.042)	&0.791(0.058)\\
		&DiMSC		&0.731(0.016)	&0.755(0.015)	&0.827(0.019)	&0.661(0.018)	&0.743(0.016)	&0.751(0.021)\\
		&RMSC		& 0.677(0.029)	&0.668(0.031)	&0.686(0.026)	&0.682(0.020)	&0.624(0.034)	&0.804(0.026)\\	
		&LMSC 		&0.656(0.081)	&0.646(0.081)	&0.667(0.080)	&0.677(0.065)	&0.600(0.094)	&0.783(0.086)\\
		&ECMSC		&0.879(0.007)	&0.876(0.008)	&0.883(0.006)	&0.870(0.005)	&0.860(0.008)	&0.937(0.001)\\
		&LTMSC  		&0.727(0.001)	&0.714(0.001)	&0.742(0.000)	&0.750(0.000)	&0.682(0.001)	&0.829(0.000)\\
		&t-SVD-MSC 	&0.962(0.000)	&0.961(0.000)	&0.963(0.000)	&0.960(0.000)	&0.955(0.000)	&0.981(0.000)\\
		&ETLMSC	&0.934(0.079)	&0.924(0.099)	&0.946(0.056)	&0.946(0.055)	&0.923(0.092)	&0.950(0.077)\\
		&HLR-M$^{2}$VS	&\underline{0.990(0.000)}	&\underline{0.990(0.000)}	&\underline{0.990(0.000)}	&\underline{0.989(0.000)}	&\underline{0.989(0.000)}	&\underline{0.995(0.000)}\\
		&CoMSC &0.861(0.000)	&0.856(0.000)	&0.961(0.000)	&863(0.000)	&838(0.000)	&0.929(0.000)\\
		&JSMC &0.667(0.000)	&0.652(0.000)	&0.682(0.000)	&0.692(0.000)	&0.612(0.000)	&0.762(0.000)\\
		&ATTN-MSC 	& \textbf{1.000(0.000)} &\textbf{1.000(0.000)}	 &\textbf{1.000(0.000)} &\textbf{1.000(0.000)} &\textbf{1.000(0.000)} &\textbf{1.000(0.000)}\\
\bottomrule
\end{tabular}
}
\end{center}
\end{table*}


\end{document}